\documentclass{article} 
\usepackage{iclr2027_conference,times}

\usepackage{amsmath,amsfonts,bm}

\def\eqref#1{equation~\ref{#1}}

\def\1{\bm{1}}

\DeclareMathAlphabet{\mathsfit}{\encodingdefault}{\sfdefault}{m}{sl}
\SetMathAlphabet{\mathsfit}{bold}{\encodingdefault}{\sfdefault}{bx}{n}

\usepackage{hyperref}
\usepackage{url}
\usepackage{booktabs}
\usepackage{amsfonts}
\usepackage{nicefrac}
\usepackage{microtype}
\usepackage{xcolor}
\usepackage{graphicx}
\usepackage{amsmath}
\usepackage{multirow}
\usepackage{subcaption}
\usepackage{tcolorbox}
\tcbuselibrary{skins, breakable}

\newtcolorbox{rqbox}[1][]{
    enhanced,
    colback=gray!3,
    colframe=black,
    fonttitle=\bfseries,
    boxed title style={
        colback=black,
        sharp corners
    },
    attach boxed title to top left={
        xshift=3mm,
        yshift=-2.5mm
    },
    title={#1}
}

\title{FSAN: Flow State Attention Network \\ for Aerodynamic Prediction}

\author{
Wenxuan Jin\textsuperscript{1,2},
Jianguo Yao\textsuperscript{1,2},
Haibing Guan\textsuperscript{1,2},
Xijun Li\textsuperscript{1,2}\thanks{Corresponding author.}\\[2mm]
\textsuperscript{1}Shanghai Key Laboratory of Scalable Computing and Systems\\
\textsuperscript{2}School of Computer Science, Shanghai Jiao Tong University
}

\iclrfinalcopy

\begin{document}

\maketitle

\begin{abstract}
Accurate aerodynamic prediction is critical for designing fuel-efficient and safe transportation systems such as aircraft and automobiles, yet traditional computational fluid dynamics (CFD) simulations remain computationally expensive and expertise-intensive, severely limiting their use in iterative design and real-time analysis. Existing deep learning surrogates suffer from two major limitations: (i) they are evaluated on datasets with narrow flow-condition ranges, leaving their performance under complex flow conditions undemonstrated; (ii) they treat global flow conditions as a single vector injected uniformly across all surface points, ignoring that different geometric regions experience distinct local flow phenomena, which degrades prediction accuracy under complex flow conditions. To address these limitations, we propose the Flow State Attention Network (FSAN). FSAN separately encodes point cloud and flow conditions, then partitions the geometry into multiple flow states via learnable soft assignments, and uses flow features to update these state representations, which in turn influence point cloud features through state changes. This enables fine-grained, state-specific interaction between geometry and flow information. Extensive experiments on two well-recognized aerodynamic benchmarks demonstrate that FSAN achieves the highest accuracy among the methods compared in this work at a higher computational cost. On Emmi-Wing, FSAN reduces the Relative L2 (REL-L2) error by over 20\% compared to the strongest baseline (Transolver), and on DrivAerNet++, it achieves a 10\% reduction compared to the strongest baseline (AdaField). These results establish FSAN as a promising neural surrogate on public benchmarks with diverse flow conditions and geometries. The code is available in the supplementary material.
\end{abstract}

\section{Introduction}
Aerodynamic performance is a critical factor in the design of transportation systems such as aircraft, automobiles, and trains. It directly influences fuel efficiency, operational range and overall safety, making accurate prediction a fundamental requirement in engineering practice. Traditional methods rely on computational fluid dynamics (CFD) simulations, which solve the Navier–Stokes equations at high fidelity \citep{anderson2023fundamentals}. However, CFD simulations are computationally expensive, requiring hours to days per configuration, and demand substantial hardware resources. These limitations restrict the application of CFD in iterative design optimization and real-time analysis.

In recent years, deep learning has emerged as a promising alternative for aerodynamic prediction. Neural surrogate models learn a direct mapping from geometry and flow conditions to aerodynamic coefficients. Consequently, they can produce predictions within seconds without requiring high-performance computing, effectively addressing the computational barriers of traditional CFD \citep{secco2017artificial, yetkin2024investigation, sung2025blendednet}. Architectures originally developed for 3D point cloud understanding, such as PointNet \citep{qi2017pointnet} and its successors \citep{qi2017pointnet++}, were subsequently adopted by the CFD community for aerodynamic prediction tasks, demonstrating that point-based deep learning could serve as an effective surrogate for traditional simulations. Graph neural networks \citep{pfaff2020learning} were introduced to better capture the connectivity of mesh-based representations, and transformer-based architectures such as PointTransformer \citep{PointTransformer} and Transolver \citep{wu2024transolver} were developed to model long-range dependencies and physical correlations. However, these existing approaches are constrained by two major limitations. First, \textit{they are evaluated on datasets where flow conditions vary only within narrow ranges}, thus their performance under complex conditions has not been demonstrated. Second, \textit{they lack a mechanism for spatially varying conditioning}, as they typically encode flow conditions as a single global vector and inject it uniformly across all surface points, ignoring that different regions of a geometry may experience distinct local flow phenomena such as laminar, turbulent, attached, or separated flow. Cross-domain studies have shown that spatially varying conditioning can be advantageous: in generative modeling, SPADE \citep{park2019semantic} preserves semantic information more effectively than uniform normalization; in deformable medical image registration, CSAIN \citep{wang2023conditional} improves local control and registration accuracy compared to a single global regularization hyperparameter.

\begin{figure}[t]
    \centering
    \includegraphics[width=\linewidth]{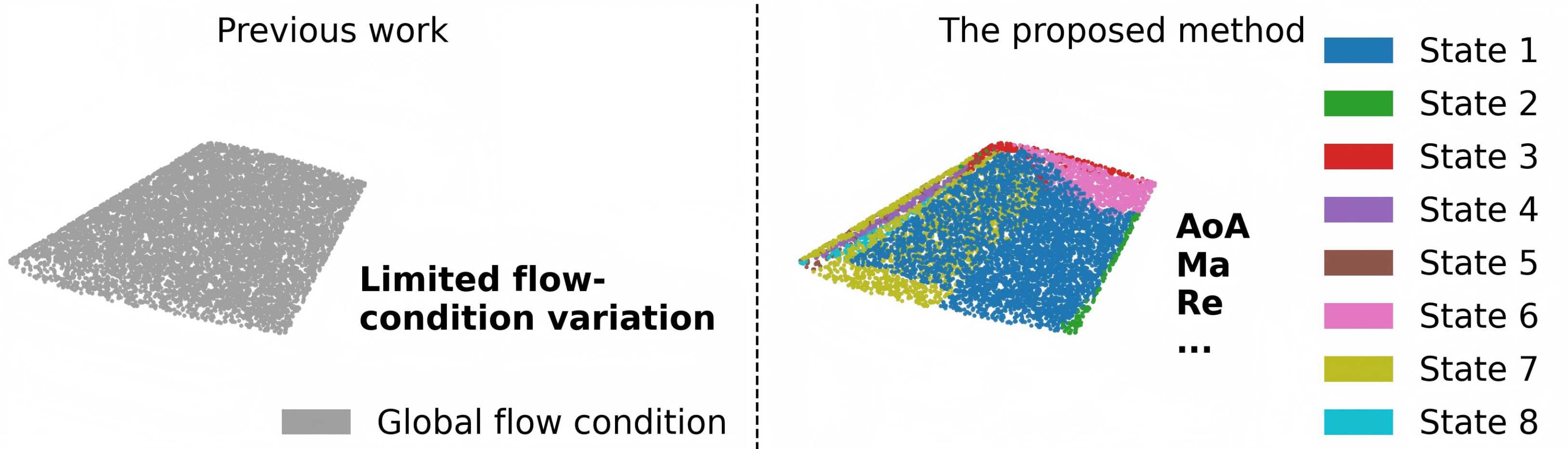}
    \caption{Conceptual comparison between existing methods and our proposed FSAN. \textbf{Left:} Many existing methods typically evaluate under limited flow-condition variation and apply a uniform global conditioning vector to all surface points. \textbf{Right:} FSAN is designed for diverse flow conditions and adaptively partitions the geometry into multiple latent states.}
    \label{fig:idea_comparison}
\end{figure}

To address these limitations, we propose the Flow State Attention Network (FSAN), which is capable of capturing region-dependent responses to flow conditions. As illustrated in Figure~\ref{fig:idea_comparison}, FSAN learns to partition the point cloud into multiple flow states and performs fine-grained, state-specific interaction between geometry and flow information. ``Flow state'' is not a predefined aerodynamic regime, but a data-driven latent prototype. Technically, FSAN encodes the point cloud and the flow conditions separately, and then partitions them into multiple flow states to perform fine-grained information fusion via flow state attention. Flow state attention uses flow features to update flow state features, and then influences point cloud features through the changes in flow states. This approach enables the point cloud to interact with multiple flow-conditioned latent partitions rather than a single, spatially uniform flow representation. Our method outperforms the compared baselines while using the fewest parameters on both DrivAerNet++ \citep{NEURIPS2024_013cf29a} and Emmi-Wing \citep{paischer2025going}, with moderate computational cost. These results establish FSAN as a promising neural surrogate on public aerodynamic benchmarks.

The main contributions in this work are as follows:

\begin{itemize}
    \item We propose FSAN, a model for accurate aerodynamic prediction under complex flow conditions. To overcome the limitation of uniform global conditioning in existing methods, we design a flow state attention mechanism that enables fine-grained, state-specific interaction between geometry and flow information.
    \item Compared to the evaluated baselines, FSAN achieves superior accuracy with substantially fewer parameters. On Emmi-Wing, it reduces the REL-L2 error by over 20\% relative to the strongest baseline (Transolver) while using about 14\% fewer parameters; on DrivAerNet++, it reduces the REL-L2 error by 10\% relative to the strongest baseline (AdaField) while using about 94\% fewer parameters.
    \item FSAN qualitatively shows spatially coherent latent partitions that are progressively refined by flow information without explicit physical supervision. Different regions of the geometry are assigned to distinct states that adapt to varying flow conditions, forming data-driven, emergent patterns aligned with geometric and flow features.
\end{itemize}

\section{Related Work}
\label{sec:related}
\subsection{Geometry-Centric Aerodynamic Prediction}
Many neural surrogates focus on capturing geometric features from point clouds, with architectures such as FIGConvNet~\citep{choy2025factorized}, DoMINO~\citep{ranade2025domino}, and TripNet~\citep{chen2025tripnet} achieving strong results via factorized grids, local geometric operators, and triplane encodings, respectively. However, these methods primarily handle geometric variation, while flow conditions are often kept within narrow ranges. Widely used benchmarks reflect this bias: DrivAerNet++~\citep{NEURIPS2024_013cf29a} uses a single freestream velocity, FlowBench~\citep{tali2024flowbench} varies only the Reynolds number, and BlendedNet~\citep{sung2025blendednet} offers only a small number of flight conditions. To assess model performance under complex, multi-parameter flow regimes, we evaluate our method and baselines on Emmi-Wing~\citep{paischer2025going}, which provides wide ranges of angle of attack, Mach number, and Reynolds number.

\subsection{Homogeneous Flow Conditioning}
Effectively integrating global flow conditions with geometry remains challenging. Simple concatenation is often insufficient, motivating feature modulation (e.g., FiLM~\citep{perez2018film}) that applies learned affine transformations to point features, and cross-attention mechanisms~\citep{Aerodynamic-Cross-Attention} that let geometric tokens attend to flow tokens. Despite their flexibility, these conditioning strategies are spatially uniform within each sample: FiLM shares the same affine parameters across all points, and cross-attention exposes every point to the same set of flow tokens. Consequently, while they can adapt to different global conditions, they do not explicitly provide a representation capable of capturing region-dependent responses that may arise from distinct local flow phenomena---such as attached flow, turbulent boundary layers, or shock-induced separation---that coexist on a single surface. We address this gap with flow state attention, which learns to partition the point cloud into multiple states and applies state-specific conditioning, enabling fine-grained, region-dependent interaction between geometry and flow conditions.

\section{Method}
\label{sec:method}

\begin{figure}
  \centering
  \includegraphics[width=\linewidth]{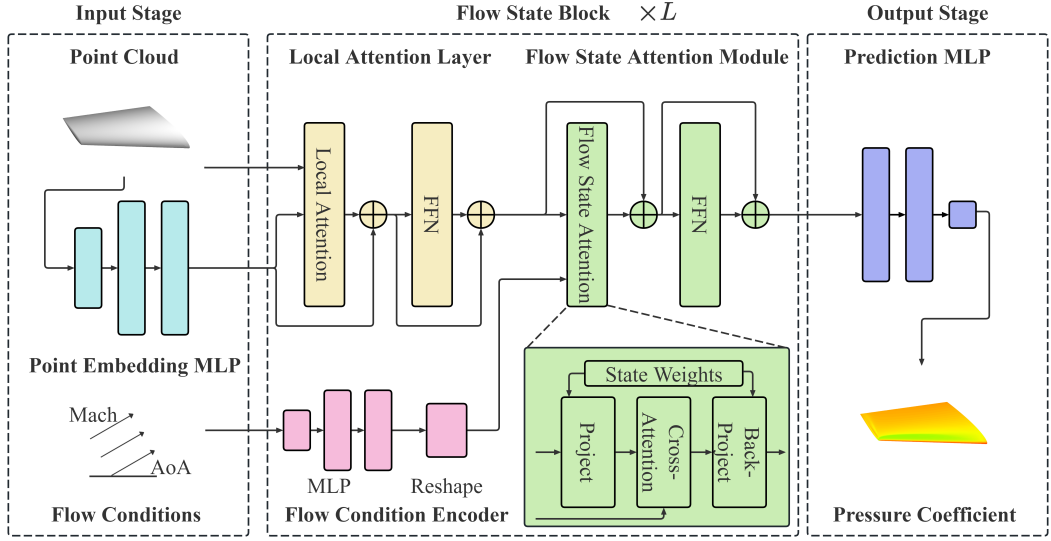}
\caption{Overall architecture of FSAN. The input point cloud is first projected to initial features by a point embedding MLP. These features, together with the original point coordinates and flow conditions, are then processed by $L$ stacked blocks. Each block consists of three modules: (i) a local attention layer that updates point features by attending to $k$ nearest neighbors; (ii) a flow condition encoder that maps flow conditions to flow features; (iii) a flow state attention module that learns a soft assignment of points to flow states, aggregates point features into state representations, updates these states via cross-attention with the flow features, and finally projects the refined states back to individual points. Finally, a prediction MLP outputs the pressure coefficient for each point.}
  \label{architecture}
\end{figure}

As discussed in Section~\ref{sec:related}, existing deep learning surrogates for aerodynamic prediction suffer from two major limitations: they focus predominantly on geometric shape variations while flow conditions are often kept within narrow ranges, and they typically encode global flow conditions as a single vector injected uniformly across all surface points, ignoring distinct local flow states. To address both limitations, we propose FSAN, which explicitly models the interaction between geometry and flow conditions by adaptively partitioning the point cloud into multiple flow states and using these states to update point features. Through the soft assignment mechanism, each point is routed to multiple such prototypes, enabling the network to decompose a complex flow field into a set of learnable conditioning bases without any explicit physical supervision. These prototypes are distributed spatially, and the points assigned to each prototype tend to form coherent regions that reflect the model's learned decomposition of the flow–geometry interaction. This section describes the FSAN architecture; training details are provided in the Appendix~\ref{app:training}.

Since surface pressure coefficient ($C_p$) fields are directly linked to integral aerodynamic quantities such as lift ($C_L$) and drag ($C_D$) through surface integration, accurate $C_p$ prediction serves as a representative and physically meaningful surrogate task for aerodynamic performance evaluation. We therefore focus on predicting the per-point $C_p$ distribution in this work.

The architecture of FSAN is illustrated in Figure~\ref{architecture}. The input of the network comprises two parts: the point cloud and the flow conditions. The point cloud is denoted as $\mathbf{P} = [\mathbf{p}_1, \mathbf{p}_2, \dots, \mathbf{p}_N]^\top \in \mathbb{R}^{N\times3}$, where $\mathbf{p}_i \in \mathbb{R}^3$ represents the Cartesian coordinates of the $i$-th surface point. The flow conditions are given by a vector $\mathbf{f} \in \mathbb{R}^{d}$ that captures the global flow parameters, where $d$ is the number of parameters. For aircraft, $\mathbf{f}$ consists of the angle of attack (AoA), Mach number, and Reynolds number. The output of the network is a vector $\hat{\mathbf{y}} \in \mathbb{R}^N$, where each component $\hat{y}_i = \hat{C}_{p,i}$ denotes the predicted pressure coefficient at the corresponding point $\mathbf{p}_i$.

FSAN adopts a stacked architecture consisting of $L$ blocks, each comprising three modules: a local attention layer, a flow condition encoder, and a flow state attention module. The local attention layer takes the point features from the previous block and updates them via local attention. The flow condition encoder maps the global flow conditions to flow features. The flow state attention module first partitions the point cloud features into multiple flow states, then performs cross‑attention between these flow states and the flow features, and finally projects the updated flow state features back to each point. The output of each block serves as the input to the next block. Before the first block, a point embedding MLP projects the point coordinates into initial point features. After the final block, the resulting point features are fed into a prediction MLP to predict the final pressure coefficient for each point.

\subsection{Point embedding MLP}

The point coordinates $\mathbf{p}_i \in \mathbb{R}^3$ are first transformed into initial point features by a point embedding MLP. This MLP independently maps each coordinate to a $C$-dimensional vector:
\begin{equation}
\mathbf{x}_i^{(0)} = \mathrm{MLP}_{\text{embed}}(\mathbf{p}_i) \in \mathbb{R}^C, \quad i = 1,\dots,N.
\end{equation}
The resulting initial features $\mathbf{X}^{(0)} = [\mathbf{x}_1^{(0)}, \dots, \mathbf{x}_N^{(0)}]^\top \in \mathbb{R}^{N \times C}$ are then fed into the first block of FSAN.

\subsection{Local attention layer}
The local attention layer adopts the local attention mechanism from the original PointTransformer \citep{PointTransformer} to enable each point to interact with its neighboring points. Specifically, we employ K-nearest neighbors (KNN) \citep{KNN} to identify the neighboring points for each point, followed by local attention computation between each point and its neighbors. For simplicity, we adopt conventional scalar attention.

The $l$-th local attention layer takes two inputs: the point features from the previous block, denoted as $\mathbf{X}^{(l-1)} = [\mathbf{x}_1^{(l-1)}, \dots, \mathbf{x}_N^{(l-1)}]^\top \in \mathbb{R}^{N \times C}$, and the point coordinates $\mathbf{P} = [\mathbf{p}_1, \dots, \mathbf{p}_N]^\top \in \mathbb{R}^{N \times 3}$, which remain fixed across all layers. For brevity, we write $\mathbf{x}_i = \mathbf{x}_i^{(l-1)}$ for the input feature of point $i$. Two MLPs are introduced: $\theta: \mathbb{R}^3 \to \mathbb{R}^C$ encodes the relative position between two points, and $\gamma: \mathbb{R}^C \to \mathbb{R}^C$ transforms the sum of the key and the positional encoding into a vector for computing the attention logit. For each point $i$, we first identify its $k$ nearest neighbors in the coordinate space, denoted by $\mathcal{N}(i) = \mathrm{KNN}(\mathbf{p}_i, \mathbf{P}, k)$. The local attention mechanism then computes the updated feature $\mathbf{x}_i'$ as follows:
\begin{equation}
\begin{aligned}
\mathbf{q}_i &= \mathbf{W}_Q \mathbf{x}_i,\quad
\mathbf{k}_{j} = \mathbf{W}_K \mathbf{x}_j,\quad
\mathbf{v}_{j} = \mathbf{W}_V \mathbf{x}_j, \\
\boldsymbol{\delta}_{ij} &= \theta(\mathbf{p}_j - \mathbf{p}_i), \\
e_{ij} &= \mathbf{q}_i^\top \bigl( \gamma(\mathbf{k}_{j} + \boldsymbol{\delta}_{ij}) \bigr), \\
\alpha_{ij} &= \frac{\exp(e_{ij})}{\sum_{j' \in \mathcal{N}(i)} \exp(e_{ij'})}, \\
\mathbf{x}_i' &= \sum_{j \in \mathcal{N}(i)} \alpha_{ij} \bigl( \mathbf{v}_{j} + \boldsymbol{\delta}_{ij} \bigr),
\end{aligned}
\end{equation}
where $\mathbf{W}_Q, \mathbf{W}_K, \mathbf{W}_V \in \mathbb{R}^{C \times C}$ are learnable linear projections. The resulting $\mathbf{x}_i'$ is then combined with the input feature $\mathbf{x}_i$ via a residual connection, followed by a feed‑forward network (FFN), yielding the output of the local attention layer, which we denote as $\tilde{\mathbf{x}}_i$:
\begin{equation}\label{eq:pt_output}
\begin{aligned}
\mathbf{x}_i'' &= \mathrm{LN}(\mathbf{x}_i' + \mathbf{x}_i), \\
\tilde{\mathbf{x}}_i &= \mathrm{LN}\bigl(\mathbf{x}_i'' + \mathrm{FFN}(\mathbf{x}_i'')\bigr),
\end{aligned}
\end{equation}
where $\mathrm{LN}(\cdot)$ denotes Layer Normalization.

\subsection{Flow condition encoder}
The flow condition vector $\mathbf{f} \in \mathbb{R}^{d}$ (with $d=3$ for aircraft tasks, including AoA, Mach number, and Reynolds number) is first encoded by an MLP into a high‑dimensional vector $\mathbf{h} \in \mathbb{R}^{K \cdot C}$. Here $C$ denotes the hidden feature dimension. To facilitate subsequent operations, this vector is reshaped into a matrix $\mathbf{F}_{\text{flow}} \in \mathbb{R}^{K \times C}$, which can be viewed as a set of $K$ learnable tokens each of dimension $C$. The transformation is formalized as:
\begin{equation}
\begin{aligned}
\mathbf{h} &= \mathrm{MLP}_{\text{flow}}(\mathbf{f}), \\ 
\mathbf{F}_{\text{flow}} &= \operatorname{reshape}\bigl( \mathbf{h}, \, K, \, C \bigr). 
\end{aligned}
\end{equation}

\subsection{Flow state attention module}

\begin{figure}
  \centering
  \includegraphics[width=\linewidth]{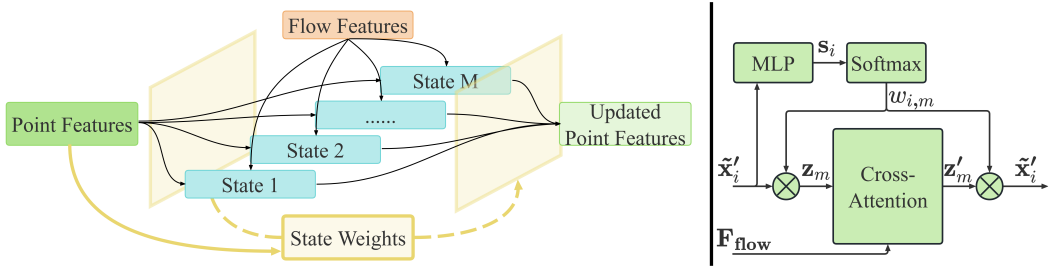}
\caption{Illustration of the flow state attention module. Given the point features $\tilde{\mathbf{x}}_i$ from the local attention layer in the same block, the module first computes state weights $\mathbf{s}_i$ via a learnable MLP, followed by a softmax to obtain normalized weights $w_{i,m}$ ($m=1,\dots,M$). These weights are used to aggregate point features into $M$ distinct flow state features $\mathbf{z}_m$ (via weighted averaging). Each flow state is then updated by cross-attention that uses the state features as queries and the global flow tokens $\mathbf{F}_{\text{flow}}$ as keys and values, yielding refined state features $\mathbf{z}_m'$. Finally, the updated state features are projected back to each point using the same state weights, producing the output point features $\tilde{\mathbf{x}}_i'$.}
  \label{flow_state_attention}
\end{figure}

The flow state attention module draws inspiration from the physics-attention mechanism of Transolver \citep{wu2024transolver}. As illustrated in Figure~\ref{flow_state_attention}, we first compute state weights from point features, then distribute the points into distinct flow states according to these weights. Each flow state is updated via cross‑attention with the global flow conditions, and finally the updated states are projected back to individual points using the same weights.

Specifically, we first compute the state weights from each point to each flow state. For point $i$, let $\tilde{\mathbf{x}}_i$ denote the point feature produced by the local attention layer in the same block. A raw score vector $\mathbf{s}_i \in \mathbb{R}^{M}$ ($M$ is the number of flow states) is produced by a learnable mapping, and the softmax operation yields normalized weights $w_{i,m}$:
\begin{equation}
\begin{aligned}
\mathbf{s}_i &= \mathrm{MLP}_{\text{weight}}(\tilde{\mathbf{x}}_i), \\
w_{i,m} &= \mathrm{Softmax}(\mathbf{s}_i)_m, \qquad m = 1,\dots,M,
\end{aligned}
\end{equation}
where $\mathrm{MLP}_{\text{weight}}$ maps the point feature to logits. Each flow state is then represented as a weighted aggregation of point features:
\begin{equation}
\mathbf{z}_m = \frac{\sum_{i=1}^{N} w_{i,m} \tilde{\mathbf{x}}_i}{\sum_{i=1}^{N} w_{i,m}},
\end{equation}
with $\mathbf{z}_m \in \mathbb{R}^{C}$ being the feature of the $m$-th flow state. These state features are subsequently updated by interacting with the global flow condition tokens $\mathbf{F}_{\text{flow}} \in \mathbb{R}^{K \times C}$ through multi-head cross‑attention:
\begin{equation}
\mathbf{z}_m' = \mathrm{CrossAttn}(\mathbf{z}_m, \mathbf{F}_{\text{flow}}), 
\end{equation}
where $\mathrm{CrossAttn}$ uses $\mathbf{z}_m$ as query and $\mathbf{F}_{\text{flow}}$ as key/value. Finally, the updated point features $\tilde{\mathbf{x}}_i'$ are obtained by projecting the refined flow states back to each point using the same weights:
\begin{equation}
\tilde{\mathbf{x}}_i' = \sum_{m=1}^{M} w_{i,m} \mathbf{z}_m'.
\end{equation}

The updated point features $\tilde{\mathbf{x}}_i'$ are then combined with the input $\tilde{\mathbf{x}}_i$ via a residual connection, followed by an FFN identical to Eq.~\eqref{eq:pt_output}. This yields the final output of the current block, denoted as $\mathbf{x}_i^{(l)}$:
\begin{equation}\label{eq:fsa_output}
\begin{aligned}
\tilde{\mathbf{x}}_i'' &= \mathrm{LN}(\tilde{\mathbf{x}}_i' + \tilde{\mathbf{x}}_i), \\
\mathbf{x}_i^{(l)} &= \mathrm{LN}\bigl(\tilde{\mathbf{x}}_i'' + \mathrm{FFN}(\tilde{\mathbf{x}}_i'')\bigr).
\end{aligned}
\end{equation}
The resulting $\mathbf{X}^{(l)} = [\mathbf{x}_1^{(l)}, \dots, \mathbf{x}_N^{(l)}]^\top \in \mathbb{R}^{N \times C}$ is passed to the next block.

In the first block, the state assignments are derived solely from the point features output by the local attention layer, thus providing a geometry-driven initial partition of the surface. Once the states are updated by cross-attention with the flow tokens in that block, the resulting point features carry flow-conditioned information. Consequently, in all subsequent blocks, the state assignments are jointly determined by geometry and the evolving flow-aware representations. This layered design allows FSAN to start from a purely geometric decomposition and progressively refine the partition under the influence of the inflow parameters.

Compared to the physics-attention mechanism in Transolver that inspired this design, FSAN differs in two fundamental aspects. First, the two models organize geometry and flow information differently. Transolver concatenates coordinates with flow conditions at the input and computes slice assignments from features that intermix both sources throughout. As described above, FSAN instead separates these streams: the initial assignment is geometry-driven, and flow conditions are introduced through cross-attention, so the partition is progressively refined across blocks rather than being passively mixed from the outset. Second, the grouped representations interact differently with the network. In Transolver, slices interact among themselves through self-attention. In FSAN, each state independently attends to the flow tokens via cross-attention, enabling different states to draw different information from the same global flow condition. The flow-conditioned states are then projected back to individual points, so that points assigned to different states receive different conditioning signals even under identical inflow parameters.

\subsection{Prediction MLP}
After the final block (the $L$-th block), the output point features $\mathbf{x}_i^{(L)} \in \mathbb{R}^C$ are passed to a prediction MLP to obtain the pressure coefficient for each point:
\begin{equation}
\hat{C}_{p,i} = \mathrm{MLP}_{\text{pred}}(\mathbf{x}_i^{(L)}) \in \mathbb{R}, \quad i = 1,\dots,N.
\end{equation}
The collection of predictions forms the output vector $\hat{\mathbf{y}} = [\hat{C}_{p,1}, \dots, \hat{C}_{p,N}]^\top \in \mathbb{R}^N$.

\section{Experiments}
\label{sec:exp}
To comprehensively evaluate our proposed FSAN, we conduct extensive experiments on two public aerodynamic datasets from the aircraft and automotive domains. We benchmark FSAN against three strong baselines. We aim to answer the following research questions:

\begin{rqbox}[Research Questions]
\textbf{RQ1:} How does FSAN compare to SOTA baselines in terms of both prediction accuracy and computational cost? (answered in Section~\ref{sec:main_results})

\textbf{RQ2:} What is the isolated contribution of the flow state attention mechanism compared with other conditioning modules on the same backbone? (Section~\ref{sec:mechanism_isolation})

\textbf{RQ3:} Does FSAN's performance scale favorably with model size? (answered in Section~\ref{sec:model_size})

\textbf{RQ4:} What additional analyses are provided to further validate FSAN? (see Appendix~\ref{app:additional_results})
\end{rqbox}

\subsection{Experimental Setup}
We compare FSAN against three representative baselines: PointTransformerV3~\citep{wu2024point}, Transolver~\citep{wu2024transolver}, and AdaField~\citep{zou2026adafieldgeneralizablesurfacepressure}. To ensure a fair comparison in model capacity, we increase the parameter count of Transolver to 18M while preserving its core mechanism. PointTransformerV3 and AdaField use their default configurations, containing 46M and 250M parameters, respectively. In contrast, our FSAN contains only 16M parameters. All models were trained under identical hyperparameters without model-specific tuning.

We evaluate on two public datasets: Emmi-Wing~\citep{paischer2025going} for aircraft aerodynamics and DrivAerNet++~\citep{NEURIPS2024_013cf29a} for automotive aerodynamics. Emmi-Wing provides variable multi-parameter flow conditions (angle of attack, Mach number, Reynolds number), while DrivAerNet++ offers wide geometric diversity under a fixed velocity. We use the PIDA augmentation~\citep{zou2026adafieldgeneralizablesurfacepressure} on DrivAerNet++ to introduce variable velocity as a flow condition, making the task more complex and enabling a more comprehensive evaluation.

We evaluate model performance using four standard regression metrics: Mean Squared Error (MSE), Mean Absolute Error (MAE), Relative L1 Error (REL-L1), and Relative L2 Error (REL-L2). For all metrics, lower values indicate better predictive accuracy. 

Detailed model configurations and baseline selection rationale are provided
in Appendix~\ref{app:model_configs}. Dataset descriptions and dataset
selection rationale are provided in Appendix~\ref{app:dataset}. Training
settings and metric definitions are provided in Appendices~\ref{app:training}
and \ref{app:metrics}.

\subsection{Main Results}
\label{sec:main_results}

\begin{table}
\small
\setlength{\tabcolsep}{4pt}
\caption{Quantitative comparison on Emmi-Wing and DrivAerNet++ datasets. Best results are in \textbf{bold}. All values are reported as mean $\pm$ standard deviation over three independent training runs with different random seeds. }
\label{tab:main_results}
\centering
\begin{tabular}{@{}l l cccc@{}}
\toprule
\multirow{2}{*}{Dataset} & \multirow{2}{*}{Method} & \multicolumn{1}{c}{MSE$\downarrow$} & \multicolumn{1}{c}{MAE$\downarrow$} & \multicolumn{1}{c}{REL-L1$\downarrow$} & \multicolumn{1}{c}{REL-L2$\downarrow$} \\
 & & ($\times 10^{-3}$) & ($\times 10^{-2}$) & (\%) & (\%) \\
\midrule
\multirow{4}{*}{Emmi-Wing}
 & PointTransformerV3 & $2.722 \pm 0.008$ & $1.637 \pm 0.005$ & $5.17 \pm 0.02$ & $11.65 \pm 0.01$ \\
 & Transolver          & $0.375 \pm 0.003$ & $0.517 \pm 0.001$ & $1.58 \pm 0.00$ & $3.24 \pm 0.03$ \\
 & AdaField            & $0.439 \pm 0.005$ & $0.620 \pm 0.001$ & $1.92 \pm 0.00$ & $3.67 \pm 0.02$ \\
 & \textbf{FSAN (Ours)} & $\bm{0.297 \pm 0.001}$ & $\bm{0.456 \pm 0.001}$ & $\bm{1.40 \pm 0.00}$ & $\bm{2.59 \pm 0.01}$ \\
\cmidrule{1-6}
\multirow{4}{*}{DrivAerNet++}
 & PointTransformerV3 & $9.970 \pm 0.019$ & $4.547 \pm 0.002$ & $17.33 \pm 0.00$ & $29.51 \pm 0.01$ \\
 & Transolver          & $4.638 \pm 0.009$ & $3.617 \pm 0.004$ & $13.78 \pm 0.02$ & $20.06 \pm 0.03$ \\
 & AdaField            & $4.559 \pm 0.037$ & $3.701 \pm 0.002$ & $14.10 \pm 0.01$ & $19.89 \pm 0.04$ \\
 & \textbf{FSAN (Ours)} & $\bm{3.675 \pm 0.021}$ & $\bm{3.273 \pm 0.024}$ & $\bm{12.54 \pm 0.02}$ & $\bm{17.91 \pm 0.02}$ \\
\bottomrule
\end{tabular}
\end{table}

\begin{table}[t]
\small
\centering
\caption{Computational cost comparison on Emmi‑Wing.}
\label{tab:cost}
\begin{tabular}{@{}lcccccc@{}}
\toprule
Model & Params & Training Time & Inference Time & Train Peak Mem. & Infer. Peak Mem. \\
     & (M)   & (s / batch)  & (s / batch)  & (GB)            & (GB) \\
\midrule
PointTransformerV3 & 46.17 & 0.165 $\pm$ 0.004 & 0.050 $\pm$ 0.000 & 2.20 & 0.47 \\
Transolver         & 18.34 & 0.122 $\pm$ 0.002 & 0.037 $\pm$ 0.000 & 3.85 & 0.17 \\
AdaField           & 247.09 & 0.978 $\pm$ 0.021 & 0.455 $\pm$ 0.004 & 8.63 & 1.34 \\
FSAN (Ours)        & 15.74 & 0.339 $\pm$ 0.001 & 0.153 $\pm$ 0.000 & 11.53 & 0.99 \\
\bottomrule
\end{tabular}
\end{table}

Table~\ref{tab:main_results} reports the accuracy results. FSAN consistently outperforms all baselines across every metric on both datasets. On Emmi-Wing, it reduces the REL-L2 error by over 20\% compared to Transolver (2.59\% vs.\ 3.24\%) with 14\% fewer parameters; on DrivAerNet++, it achieves a 10\% reduction over AdaField (17.91\% vs.\ 19.89\%) with 94\% fewer parameters. These gains hold across all four metrics, validating the effectiveness of flow state attention under both diverse flow conditions and wide geometric variation. 

Table~\ref{tab:cost} reports the computational cost measured on Emmi-Wing under identical hardware, batch size, and point count settings. Corresponding DrivAerNet++ measurements are provided in Appendix~\ref{app:cost_drivaer}. FSAN's inference is about 4$\times$ slower than Transolver and uses about 5.8$\times$ more GPU memory. Module-level analysis (Appendix~\ref{app:module_cost}) shows that this overhead arises primarily from the non-hierarchical local attention, which retains full-resolution activations for all 8,192 points. Efficient backbone variants remain an important future direction for reducing this cost.

\subsection{Ablation Studies}
\label{sec:ablation_fsa}
We conduct ablation studies exclusively on the Emmi-Wing dataset because its multi-parameter flow conditions (angle of attack, Mach number, and Reynolds number) are essential for evaluating the flow state attention mechanism.

\subsubsection{Mechanism isolation}
\label{sec:mechanism_isolation}
We performed controlled experiments on the full FSAN backbone, replacing only the geometry–flow fusion mechanism while keeping local attention and flow condition encoder fixed. The results are in Table~\ref{tab:mechanism_isolation}. FSAN achieves the best accuracy among all tested conditioning methods. 

\begin{table}[ht]
\small
\centering
\caption{Mechanism isolation ablation on Emmi-Wing.}
\label{tab:mechanism_isolation}
\begin{tabular}{@{}lcccc@{}}
\toprule
Variant & MSE$\downarrow$ ($\times 10^{-3}$) & MAE$\downarrow$ ($\times 10^{-2}$) & REL-L1$\downarrow$ (\%) & REL-L2$\downarrow$ (\%) \\
\midrule
\textbf{FSAN (flow state attention)} & \textbf{0.297} & \textbf{0.456} & \textbf{1.40} & \textbf{2.59} \\
FSAN + FiLM & 0.334 & 0.501 & 1.65 & 2.83 \\
FSAN + AdaLN-Zero & 0.325 & 0.499 & 1.61 & 2.80 \\
FSAN + Cross-Attn & 0.374 & 0.602 & 1.87 & 3.16 \\
FSAN + Physics-Attn & 0.298 & 0.484 & 1.49 & 2.68 \\
\bottomrule
\end{tabular}
\end{table}

\subsubsection{Model size}
\label{sec:model_size}
\begin{figure}[t]
    \centering
    \includegraphics[width=0.6\linewidth]{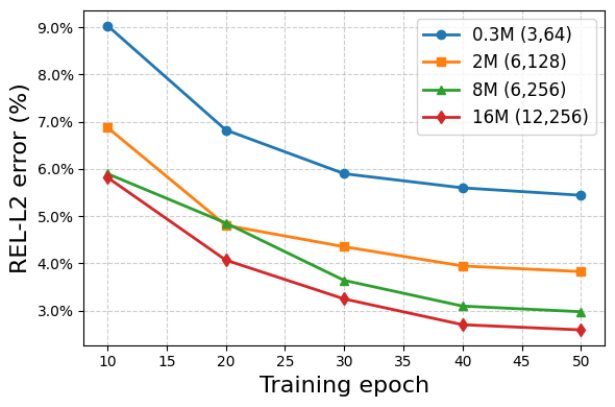}
    \caption{Training convergence of FSAN variants with different model sizes. Numbers in parentheses denote (number of blocks $L$, feature dimension $C$).}
    \label{fig:ablation_convergence}
\end{figure}

We further investigate model capacity by varying the number of blocks $L$ and the feature dimension $C$. Figure~\ref{fig:ablation_convergence} shows training convergence for four variants (0.3M, 2M, 8M, 16M parameters; $(L,C)$ values indicated in the legend). Both final error and convergence speed improve consistently with model size. The 16M configuration (12 blocks, 256 dimensions) achieves the lowest error among the tested variants and is used in our full model.

\subsection{Additional Analyses}
\label{sec:additional_analyses}
In the appendices, we provide further analyses that complement the main
results: comparison with AB‑UPT (Appendix~\ref{app:abupt}), effect of the
number of flow states (Appendix~\ref{app:num_states}), conditioning
strategy comparison on Transolver (Appendix~\ref{app:transolver_conditioning}),
computational cost analysis (Appendix~\ref{app:cost_analysis}), sensitivity
to sampling density (Appendix~\ref{app:sensitivity}), and consistency between
full‑surface and sampled predictions (Appendix~\ref{app:full_surface}).
Qualitative results, including prediction quality, evolution of flow state
assignments, and statistical association with shock regions, are presented
in Appendix~\ref{app:qualitative}.

\section{Discussion}
\label{sec:discussion}
Despite the promising results, this work has several limitations. First, FSAN is evaluated on Emmi-Wing and DrivAerNet++ separately, and its cross-domain generalization remains unexplored. Second, its performance in data-scarce scenarios, common in engineering design, has not been investigated. Third, out-of-distribution evaluation is missing, partly due to the scarcity of public OOD aerodynamic datasets. Fourth, FSAN incurs a higher computational cost than lightweight baselines such as Transolver due to its non-hierarchical design that retains full-resolution activations.

Future work will address these limitations in several directions: joint training and transfer learning for cross-domain generalization; meta-learning and multi-fidelity approaches for data scarcity; systematic OOD evaluation once suitable datasets become available; hierarchical approximations and efficient attention variants to reduce computational cost. Besides, uncertainty quantification is an important issue for safety-critical aerodynamic prediction, and we will explore this as a future research direction.

\section{Conclusion}
In this work, we proposed FSAN, which addresses two key shortcomings of existing deep learning surrogates: the lack of evaluation under complex flow conditions and the homogeneous conditioning that disregards local flow phenomena. By learning a soft, flow-conditioned partition of the geometry into multiple latent states, FSAN enables fine-grained, state-specific interaction between geometry and flow conditions. Evaluated on two public aerodynamic benchmarks, FSAN achieves the best accuracy among the methods compared in this work, while using the fewest parameters, at a higher computational cost. On Emmi-Wing, it reduces the REL-L2 error by 20\% with 14\% fewer parameters; on DrivAerNet++, it reduces the REL-L2 error by 10\% with 94\% fewer parameters. These results validate the effectiveness of the flow state attention mechanism and highlight FSAN as a promising solution for accurate and parameter-efficient aerodynamic prediction.

\subsection*{AI use statement}

In this work, we used generative AI tools for grammar checking and language
polishing, code generation and debugging, figure and table formatting,
LaTeX typesetting assistance, and literature search and summarization.
All AI-assisted content was manually reviewed and verified by the authors.
We did not use generative AI tools for the core scientific design,
theoretical derivation, or primary experimental analysis. We take full
responsibility for the final content of this work, including text, code,
claims, and artifacts produced with the aid of generative AI.

\subsection*{Reproducibility statement}
We provide anonymous source code in the supplementary material, including
the implementation of FSAN and the training/evaluation scripts. We do not
redistribute the baseline implementations since all baselines are publicly
available under their original licenses, and the specific configurations
and adaptation details used in this work are described in
Appendix~\ref{app:model_configs}. Training settings are provided in
Appendix~\ref{app:training}. All experiments use public datasets:
Emmi-Wing and DrivAerNet++.

\bibliography{reference}

@InProceedings{PointTransformer,
  author    = {Zhao, Hengshuang and Jiang, Li and Jia, Jiaya and Torr, Philip H.S. and Koltun, Vladlen},
  title     = {Point {Transformer}},
  booktitle = {Proceedings of the {IEEE}/{CVF} International Conference on Computer Vision ({ICCV})},
  month     = {October},
  year      = {2021},
  pages     = {16259--16268}
}

@inproceedings{Aerodynamic-Cross-Attention,
  title     = {Aerodynamic Coefficients Prediction via Cross-Attention Fusion and Physical-Informed Training},
  author    = {Wang, Yueqing and Zhang, Peng and Liu, Yushuang and Zhao, Jianing and Lin, Jie and Chen, Yi},
  booktitle = {Proceedings of the {AAAI} Conference on Artificial Intelligence},
  volume    = {39},
  number    = {1},
  pages     = {869--876},
  year      = {2025},
}

@inproceedings{wu2024Transolver,
  title        = {{Transolver}: A Fast Transformer Solver for {PDEs} on General Geometries},
  author       = {Haixu Wu and Huakun Luo and Haowen Wang and Jianmin Wang and Mingsheng Long},
  booktitle    = {{International Conference on Machine Learning}},
  year         = {2024}
}

@ARTICLE{KNN,
  author  = {Cover, T. and Hart, P.},
  journal = {{IEEE Transactions on Information Theory}}, 
  title   = {Nearest neighbor pattern classification}, 
  year    = {1967},
  volume  = {13},
  number  = {1},
  pages   = {21-27},
}

@inproceedings{zou2026adafieldgeneralizablesurfacepressure,
  title={Adafield: generalizable surface pressure modeling with physics-informed pre-training and flow-conditioned adaptation},
  author={Zou, Junhong and Qiu, Wei and Sun, Zhenxu and Zhang, Xiaomei and Zhang, Zhaoxiang and Zhu, Xiangyu},
  booktitle={Proceedings of the AAAI Conference on Artificial Intelligence},
  volume={40},
  number={2},
  pages={1676--1684},
  year={2026}
}

@article{paischer2025going,
  author       = {Fabian Paischer and
                  L{\'{e}}o Cotteleer and
                  Yann Dreze and
                  Richard Kurle and
                  Dylan Rubini and
                  Maurits Bleeker and
                  Tobias Kronlachner and
                  Johannes Brandstetter},
  title        = {Going with the Speed of Sound: Pushing Neural Surrogates into Highly-turbulent
                  Transonic Regimes},
  journal      = {{CoRR}},
  volume       = {abs/2511.21474},
  year         = {2025},
  eprinttype   = {arXiv},
  eprint       = {2511.21474},
}

@inproceedings{NEURIPS2024_013cf29a,
    author    = {Elrefaie, Mohamed and Morar, Florin and Dai, Angela and Ahmed, Faez},
    booktitle = {Advances in Neural Information Processing Systems},
    editor    = {A. Globerson and L. Mackey and D. Belgrave and A. Fan and U. Paquet and J. Tomczak and C. Zhang},
    pages     = {499--536},
    publisher = {Curran Associates, Inc.},
    title     = {{DrivAerNet}++: A Large-Scale Multimodal Car Dataset with Computational Fluid Dynamics Simulations and Deep Learning Benchmarks},
    volume    = {37},
    year      = {2024}
}

@book{anderson2023fundamentals,
  title     = {Fundamentals of Aerodynamics},
  author    = {Anderson, John D.},
  edition   = {7th ed.},
  year      = {2023},
  publisher = {{McGraw-Hill} {US} Higher Ed {USE}}
}

@inproceedings{sung2025blendednet,
  title     = {{BlendedNet}: A Blended Wing Body Aircraft Dataset and Surrogate Model for Aerodynamic Predictions},
  author    = {Sung, Nicholas and Spreizer, Steven and Elrefaie, Mohamed and Samuel, Kaira and Jones, Matthew C and Ahmed, Faez},
  booktitle = {{International Design Engineering Technical Conferences and Computers and Information in Engineering Conference}},
  volume    = {89237},
  pages     = {V03BT03A049},
  year      = {2025},
  organization = {American Society of Mechanical Engineers}
}

@article{yetkin2024investigation,
  title     = {Investigation on the abilities of different artificial intelligence methods to predict the aerodynamic coefficients},
  author    = {Yetkin, Sadik and Abuhanieh, Saleh and Yigit, Sahin},
  journal   = {Expert Systems with Applications},
  volume    = {237},
  pages     = {121324},
  year      = {2024},
  publisher = {Elsevier}
}

@article{secco2017artificial,
  title     = {Artificial neural networks to predict aerodynamic coefficients of transport airplanes},
  author    = {Secco, Ney Rafael and Mattos, Bento Silva de},
  journal   = {Aircraft Engineering and Aerospace Technology},
  volume    = {89},
  number    = {2},
  pages     = {211--230},
  year      = {2017},
  publisher = {Emerald Publishing Limited}
}

@inproceedings{qi2017pointnet,
  title     = {{PointNet}: Deep learning on point sets for {3D} classification and segmentation},
  author    = {Qi, Charles R and Su, Hao and Mo, Kaichun and Guibas, Leonidas J},
  booktitle = {Proceedings of the {IEEE} Conference on Computer Vision and Pattern Recognition},
  pages     = {652--660},
  year      = {2017}
}

@inproceedings{pfaff2020learning,
  title     = {Learning mesh-based simulation with graph networks},
  author    = {Pfaff, Tobias and Fortunato, Meire and Sanchez-Gonzalez, Alvaro and Battaglia, Peter},
  booktitle = {{International Conference on Learning Representations}},
  year      = {2020}
}

@article{tali2024flowbench,
  title   = {{FlowBench}: A large scale benchmark for flow simulation over complex geometries},
  author  = {Tali, Ronak and Rabeh, Ali and Yang, Cheng-Hau and Shadkhah, Mehdi and Karki, Samundra and Upadhyaya, Abhisek and Dhakshinamoorthy, Suriya and Saadati, Marjan and Sarkar, Soumik and Krishnamurthy, Adarsh and others},
  journal = {{arXiv} preprint arXiv:2409.18032},
  year    = {2024}
}

@inproceedings{perez2018film,
  title     = {{FiLM}: Visual reasoning with a general conditioning layer},
  author    = {Perez, Ethan and Strub, Florian and De Vries, Harm and Dumoulin, Vincent and Courville, Aaron},
  booktitle = {Proceedings of the {AAAI} Conference on Artificial Intelligence},
  volume    = {32},
  number    = {1},
  year      = {2018}
}

@inproceedings{wu2024point,
  title     = {Point transformer v3: Simpler faster stronger},
  author    = {Wu, Xiaoyang and Jiang, Li and Wang, Peng-Shuai and Liu, Zhijian and Liu, Xihui and Qiao, Yu and Ouyang, Wanli and He, Tong and Zhao, Hengshuang},
  booktitle = {Proceedings of the {IEEE}/{CVF} Conference on Computer Vision and Pattern Recognition},
  pages     = {4840--4851},
  year      = {2024}
}

@article{qi2017pointnet++,
  title     = {{PointNet}++: Deep hierarchical feature learning on point sets in a metric space},
  author    = {Qi, Charles Ruizhongtai and Yi, Li and Su, Hao and Guibas, Leonidas J},
  journal   = {Advances in Neural Information Processing Systems},
  volume    = {30},
  year      = {2017}
}

@article{choy2025factorized,
  title   = {Factorized implicit global convolution for automotive computational fluid dynamics prediction},
  author  = {Choy, Chris and Kamenev, Alexey and Kossaifi, Jean and Rietmann, Max and Kautz, Jan and Azizzadenesheli, Kamyar},
  journal = {{arXiv} preprint arXiv:2502.04317},
  year    = {2025}
}

@article{ranade2025domino,
  title   = {{DoMINO}: A decomposable multi-scale iterative neural operator for modeling large scale engineering simulations},
  author  = {Ranade, Rishikesh and Nabian, Mohammad Amin and Tangsali, Kaustubh and Kamenev, Alexey and Hennigh, Oliver and Cherukuri, Ram and Choudhry, Sanjay},
  journal = {{arXiv} preprint arXiv:2501.13350},
  year    = {2025}
}

@article{chen2025tripnet,
  title   = {{TripNet}: Learning large-scale high-fidelity {3D} car aerodynamics with triplane networks},
  author  = {Chen, Qian and Elrefaie, Mohamed and Dai, Angela and Ahmed, Faez},
  journal = {{arXiv} preprint arXiv:2503.17400},
  year    = {2025}
}

@article{chen2023pointgpt,
  title     = {{PointGPT}: Auto-regressively generative pre-training from point clouds},
  author    = {Chen, Guangyan and Wang, Meiling and Yang, Yi and Yu, Kai and Yuan, Li and Yue, Yufeng},
  journal   = {Advances in Neural Information Processing Systems},
  volume    = {36},
  pages     = {29667--29679},
  year      = {2023}
}

@inproceedings{liu2025aerogto,
  title     = {{AeroGTO}: An efficient graph-transformer operator for learning large-scale aerodynamics of {3D} vehicle geometries},
  author    = {Liu, Pengwei and Wang, Pengkai and Ren, Xingyu and Yuan, Hangjie and Hao, Zhongkai and Xu, Chao and Cai, Shengze and Ni, Dong},
  booktitle = {Proceedings of the {AAAI} Conference on Artificial Intelligence},
  volume    = {39},
  number    = {18},
  pages     = {18924--18932},
  year      = {2025}
}

@article{ashton2024windsorml,
  title     = {{WindsorML}: High-fidelity computational fluid dynamics dataset for automotive aerodynamics},
  author    = {Ashton, Neil and Angel, Jordan B and Ghate, Aditya S and Kenway, Gaetan K and Wong, Man L and Kiris, Cetin and Walle, Astrid and Maddix, Danielle C and Page, Gary},
  journal   = {Advances in Neural Information Processing Systems},
  volume    = {37},
  pages     = {37823--37835},
  year      = {2024}
}

@article{ashton2024drivaerml,
  title   = {{DrivAerML}: High-fidelity computational fluid dynamics dataset for road-car external aerodynamics},
  author  = {Ashton, Neil and Mockett, Charles and Fuchs, Marian and Fliessbach, Louis and Hetmann, Hendrik and Knacke, Thilo and Schonwald, Norbert and Skaperdas, Vangelis and Fotiadis, Grigoris and Walle, Astrid and others},
  journal = {{arXiv} preprint arXiv:2408.11969},
  year    = {2024}
}

@article{bonnet2022airfrans,
  title     = {{AirfRANS}: High fidelity computational fluid dynamics dataset for approximating {Reynolds}-averaged {Navier}--{Stokes} solutions},
  author    = {Bonnet, Florent and Mazari, Jocelyn and Cinnella, Paola and Gallinari, Patrick},
  journal   = {Advances in Neural Information Processing Systems},
  volume    = {35},
  pages     = {23463--23478},
  year      = {2022}
}

@article{su2023awsd,
  title     = {{AWSD}: An Aircraft Wing Dataset Created by an Automatic Workflow for Data Mining in Geometric Processing},
  author    = {Su, Xiang and Li, Nan and Hu, Yuedi and Li, Haisheng},
  journal   = {Computer Modeling in Engineering \& Sciences},
  volume    = {136},
  number    = {3},
  pages     = {2935},
  year      = {2023},
  publisher = {Tech Science Press}
}

@inproceedings{park2019semantic,
  title={Semantic image synthesis with spatially-adaptive normalization},
  author={Park, Taesung and Liu, Ming-Yu and Wang, Ting-Chun and Zhu, Jun-Yan},
  booktitle={2019 IEEE/CVF conference on computer vision and pattern recognition (CVPR)},
  pages={2332--2341},
  year={2019},
  organization={IEEE}
}

@inproceedings{wang2023conditional,
  title={Conditional deformable image registration with spatially-variant and adaptive regularization},
  author={Wang, Yinsong and Qiu, Huaqi and Qin, Chen},
  booktitle={2023 IEEE 20th International Symposium on Biomedical Imaging (ISBI)},
  pages={1--5},
  year={2023},
  organization={IEEE}
}

@article{alkin2025ab,
  title={AB-UPT: Scaling neural CFD surrogates for high-fidelity automotive aerodynamics simulations via anchored-branched universal physics transformers},
  author={Alkin, Benedikt and Bleeker, Maurits and Kurle, Richard and Kronlachner, Tobias and Sonnleitner, Reinhard and Dorfer, Matthias and Brandstetter, Johannes},
  journal={arXiv preprint arXiv:2502.09692},
  year={2025}
}
\bibliographystyle{iclr2027_conference}

\newpage
\appendix

\section{Dataset Description and Selection}
\label{app:dataset}

\subsection{Emmi-Wing}
\label{app:dataset_emmi}
The Emmi-Wing \citep{paischer2025going} dataset comprises approximately 30,000 high-fidelity Reynolds-Averaged Navier-Stokes (RANS) simulations of three-dimensional wings in the transonic regime. Wing geometries are parameterized by span, taper ratio, sweep angle, and root chord, while flow conditions vary across Mach number ($0.43$ to $0.87$), Reynolds number ($5.14\times 10^6$ to $1.95\times 10^7$), and angle of attack ($-10^\circ$ to $10^\circ$). Each sample provides surface-level flow-field data. These data enable the computation of aerodynamic performance metrics such as surface pressure coefficients $C_p$. We split the dataset into training, validation, and test sets with an 8:1:1 ratio, ensuring that the flow condition distribution is similar across the three splits. For the FSAN model, the flow condition vector $\mathbf{f} \in \mathbb{R}^{3}$ input to the model consists of these three parameters: AoA, Mach number, and Reynolds number. The Emmi-Wing dataset is distributed under the CC BY-NC 4.0 license.

\subsection{DrivAerNet++}
\label{app:dataset_drivaer}
DrivAerNet++ \citep{NEURIPS2024_013cf29a} is a large-scale, high-fidelity dataset for learning-based aerodynamic analysis of vehicles. It contains over 8,000 diverse car designs modeled with steady-state incompressible simulations at 30 m/s flow velocity. Each sample provides surface point clouds with more than 500,000 points, along with corresponding pressure fields. We adopt the official split provided in the official repository. To improve generalization across different object scales and velocities, we augment the dataset using the Physics-Informed Data Augmentation (PIDA) technique proposed in AdaField \citep{zou2026adafieldgeneralizablesurfacepressure}. The flow condition vector $\mathbf{f} \in \mathbb{R}^1$ input to FSAN contains only velocity. The DrivAerNet++ dataset is distributed under the CC BY-NC 4.0 license.

\subsection{Dataset Selection Rationale}
\label{app:dataset_selection}
\begin{table}[htbp]
\centering
\caption{Overview of representative public aerodynamic datasets and their characteristics.}
\label{tab:dataset_selection}
\begin{tabular}{@{}llllp{3.5cm}@{}}
\toprule
Dataset & Domain & 3D? & Flow Condition & Geometric Diversity \\
\midrule
DrivAerNet++ & Automotive & Yes & Fixed velocity & High \\
\citep{NEURIPS2024_013cf29a} & & & (30 m/s). & (8,000+ shapes). \\
\addlinespace
Emmi-Wing & Aircraft & Yes & Variable AoA, & Moderate \\
\citep{paischer2025going} & & & Ma, Re. & (parameterized wings). \\
\addlinespace
DrivAerML & Automotive & Yes & Variable speed & Moderate. \\
\citep{ashton2024drivaerml} & & & (20--50 m/s). & \\
\addlinespace
WindsorML & Automotive & Yes & Fixed velocity. & Moderate. \\
\citep{ashton2024windsorml} & & & & \\
\addlinespace
AirfRANS & Airfoil & No (2D) & Variable AoA, Re. & Low (2D profiles). \\
\citep{bonnet2022airfrans} & & & & \\
\addlinespace
AWSD & Aircraft & Yes & Fixed flow & Moderate \\
\citep{su2023awsd} & & & conditions. & (parameterized wings). \\
\bottomrule
\end{tabular}
\end{table}

Table~\ref{tab:dataset_selection} lists representative public aerodynamic datasets and their key characteristics. Most existing benchmarks focus on geometric variation under fixed or single-parameter flow conditions. We select DrivAerNet++~\citep{NEURIPS2024_013cf29a} for its wide geometric diversity and Emmi-Wing~\citep{paischer2025going} for its comprehensive multi-parameter flow regimes, enabling a balanced evaluation of FSAN under both types of variation.

\section{Model Configurations and Baseline Selection}
\label{app:model_configs}

\subsection{FSAN}
\label{app:fsan_config}
We configure FSAN as follows:

\begin{itemize}
    \setlength{\itemsep}{0pt}
    \item \textbf{Input representation}: The model takes a point cloud of normalized $3$D coordinates as geometric input, and a flow condition vector $\mathbf{f}\in\mathbb{R}^{d}$ as the global flow input. For aircraft tasks, $\mathbf{f}$ consists of angle of attack, Mach number, and Reynolds number ($d=3$); for the PIDA-augmented DrivAerNet++ setting, $\mathbf{f}$ contains only velocity ($d=1$).
    \item \textbf{Point embedding MLP}: The point coordinates are first projected by the point embedding MLP into initial point features of dimension $C=256$.
    \item \textbf{Stacked architecture}: FSAN consists of $L=12$ stacked blocks. Each block contains a local attention layer, a flow condition encoder, and a flow state attention module.
    \item \textbf{Local attention layer}: Each local attention layer identifies the $k=16$ nearest neighbors of each point in coordinate space and performs conventional scalar attention with learnable positional encoding $\theta(\cdot)$ and score mapping $\gamma(\cdot)$.
    \item \textbf{Flow condition encoder}: The flow condition vector is mapped by an MLP and reshaped into $K=3$ flow tokens, each of dimension $C=256$.
    \item \textbf{Flow state attention module}: The flow state attention module employs $M=8$ flow states. Point features are assigned to these states via softmax-normalized weights produced by a learnable MLP, aggregated into state features, updated through cross-attention with the flow tokens, and projected back to individual points using the same weights.
    \item \textbf{Prediction MLP}: After the final block, the prediction MLP maps each point feature to a scalar pressure coefficient $\hat{C}_{p,i}$.
\end{itemize}
The total number of trainable parameters in FSAN is approximately 16 million. The detailed architecture is described in Section~\ref{sec:method}. 

\subsection{PointTransformerV3}
\label{app:ptv3}
We adopt the official PointTransformerV3 architecture \citep{wu2024point} as one of our baselines. The model is a serialized point cloud transformer designed for large-scale 3D scene understanding. For the task of surface pressure coefficient prediction, we adapt it as follows:

\begin{itemize}
    \setlength{\itemsep}{0pt}
    \item \textbf{Input representation}: Each point is represented by a $6$-dimensional feature vector: the normalized $3$D coordinates and the three normalized flow conditions (angle of attack, Mach number, Reynolds number).
    \item \textbf{Grid size}: The voxel grid size is set to $0.01$ after normalizing the point cloud to the unit cube.
    \item \textbf{Serialization}: Four serialization orders are used: \texttt{z}, \texttt{z-trans}, \texttt{hilbert}, and \texttt{hilbert-trans}, with shuffling enabled to improve robustness.
    \item \textbf{Flash attention}: Flash attention is disabled to avoid potential memory issues with point cloud inputs.
    \item \textbf{Encoder stages}: The encoder consists of $5$ stages with depths $[2,2,2,6,2]$, output channels $[32,64,128,256,512]$, and numbers of attention heads $[2,4,8,16,32]$, respectively. Downsampling is performed with stride $2$ at each stage transition using serialized pooling.
    \item \textbf{Decoder stages}: The decoder consists of $4$ stages with depths $[2,2,2,2]$, output channels $[64,64,128,256]$, and numbers of attention heads $[4,4,8,16]$, respectively. Serialized unpooling is used to upsample features, with skip connections from the encoder.
    \item \textbf{Attention block}: Each block uses a patch size of $1024$ for serialized attention, an MLP expansion ratio of $4$, a dropout path rate of $0.3$, and pre-normalization.
    \item \textbf{Output head}: A final linear layer maps the decoder's output features (channel dimension $64$) to a single scalar representing the pressure coefficient.
\end{itemize}

The total number of trainable parameters is approximately $46$ million. The detailed architecture follows the official PointTransformerV3 implementation. The official implementation is released under the MIT License.

\subsection{Transolver}
\label{app:transolver}
We adopt Transolver \citep{wu2024transolver} as another baseline. Transolver is a transformer-based neural solver for partial differential equations on general geometries, featuring a physical attention mechanism that encodes the domain into physics-aware tokens. To ensure a fair comparison in model capacity, we increase its number of layers from the default configuration of 5 (approximately 4M parameters) to 25 (approximately 18M parameters), while preserving its core physical attention mechanism and all other architectural choices (hidden dimension 256, 8 attention heads, 32 slices, MLP ratio 4). The enlarged Transolver is trained under identical hyperparameters without model-specific tuning. We adapt it as follows:

\begin{itemize}
    \setlength{\itemsep}{0pt}
    \item \textbf{Input representation}: Each point is represented by a $6$-dimensional feature vector: the normalized $3$D coordinates and the three normalized flow conditions (angle of attack, Mach number, Reynolds number).
    \item \textbf{Hidden dimension}: The transformer hidden dimension $C$ is set to $256$.
    \item \textbf{Number of layers}: The model consists of $25$ stacked Transolver blocks.
    \item \textbf{Number of attention heads}: Each block uses $8$ attention heads.
    \item \textbf{MLP expansion ratio}: The MLP in each block has an expansion ratio of $4$, i.e., the hidden dimension of the MLP is $256 \times 4 = 1024$.
    \item \textbf{Dropout}: The dropout rate for attention and MLP is set to $0.1$.
    \item \textbf{Slice number}: The physical attention mechanism partitions the domain into $32$ slices (tokens) for capturing physical correlations.
\end{itemize}

The detailed architecture follows the official Transolver implementation. The official implementation is released under the MIT License.

\subsection{AdaField}
\label{app:adafield}
We adopt AdaField \citep{zou2026adafieldgeneralizablesurfacepressure} as another baseline. AdaField is a point transformer architecture with flow-conditioned adapters, designed for aerodynamic surface pressure prediction. For our task, we use the official implementation with the following configuration:

\begin{itemize}
    \setlength{\itemsep}{0pt}
    \item \textbf{Input representation}: The model takes normalized point coordinates ($3$D) as input; flow conditions (angle of attack, Mach number, Reynolds number) are encoded separately and injected via adaptive modulation.
    \item \textbf{Encoder depth}: The encoder consists of $5$ stages with depths $[4,4,6,12,8]$, each stage containing that many PointTransformer layers.
    \item \textbf{Encoder channels}: The channel dimensions for the $5$ stages are $[64,128,256,512,1024]$.
    \item \textbf{Downsampling}: Between encoder stages, point clouds are downsampled using slot attention to target point counts $[1024,256,64,16]$.
    \item \textbf{Decoder}: The decoder mirrors the encoder with symmetric upsampling via KNN interpolation and skip connections, using the same stage depths and channels in reverse order.
    \item \textbf{Attention mechanism}: Each PointTransformer layer uses multi-head point attention with $k=16$ nearest neighbors. The number of heads for a stage with channel dimension $C$ is set to $\max(\lfloor C/64 \rfloor, 1)$, ensuring that each head has at most $64$ dimensions.
    \item \textbf{Flow-conditioned adapters}: In each transformer layer, an adapter modulates the features using the flow condition. The adapter has a hidden dimension of $64$.
    \item \textbf{Output head}: A final linear layer projects the output features to a single scalar representing the pressure coefficient.
\end{itemize}

The total number of trainable parameters is approximately $250$ million. The detailed architecture follows the official AdaField implementation. The official implementation is released under the Apache License 2.0.

\subsection{Baseline Selection Rationale}
\label{app:baseline_selection}
Table~\ref{tab:baseline_selection} summarizes the included and excluded baseline methods and the rationale for each. Several strong methods are excluded because they either do not natively accept variable flow conditions, would require substantial architectural modifications to support Emmi-Wing's multi-parameter inputs, or their code is not publicly available. The three included baselines are adapted under a uniform protocol: for models without a native flow encoder, flow parameters are concatenated with point coordinates to form a 6-dimensional input. We acknowledge that this concatenation may cause geometry-only models to partially misinterpret flow parameters as spatial dimensions; however, it is the most straightforward and widely used strategy to equip such architectures with global conditioning, and it is applied consistently to all baselines lacking a dedicated flow encoder.In Appendix~\ref{app:transolver_conditioning}, we further compare different conditioning strategies on Transolver (coordinate concatenation, FiLM, and AdaLN-Zero) and find that concatenation performs best for this baseline on our task.

The Emmi-Wing paper \citep{paischer2025going} also adopts AB-UPT as a baseline. We reproduce and evaluate AB-UPT under a comparable protocol in Appendix~\ref{app:abupt}.

\begin{table}[h]
\centering
\caption{Summary of included and excluded baseline methods.}
\label{tab:baseline_selection}
\begin{tabular}{@{}lp{3.5cm}p{8.2cm}@{}}
\toprule
Status & Method & Rationale \\
\midrule
Included & PointTransformerV3 \citep{wu2024point} & Public implementation; flow conditions injected via coordinate concatenation (same protocol as other geometry-only models). \\
& Transolver \quad\quad\quad\quad \citep{wu2024transolver} & Public implementation; physical attention mechanism; parameter count increased to 18M for capacity matching. \\
& AdaField  \qquad\qquad\quad\citep{zou2026adafieldgeneralizablesurfacepressure} & Public implementation; natively supports flow-conditioned adapters for surface pressure prediction. \\
\midrule
Excluded & FIGConvNet \qquad\quad \citep{choy2025factorized} & No native flow condition input; concatenation would cause spatial misreading of physical parameters. \\
& DoMINO  \quad\quad\quad  \citep{ranade2025domino} & Single-parameter design (velocity only); extending to multi-parameter conditions (AoA, Ma, Re) would require architectural modification of the parameter encoder. \\
& TripNet  \qquad\qquad\qquad \citep{chen2025tripnet} & Code not publicly available; triplane encoding not validated under multi-parameter flow conditions. \\
& AeroGTO  \qquad\qquad\qquad \citep{liu2025aerogto} & Graph construction over point coordinates; flow parameter concatenation would distort graph structure. \\
& PointGPT   \qquad\qquad\qquad \citep{chen2023pointgpt} & Generative pre-training model; requires extensive adaptation for regression tasks. \\
& RegDGCNN  \qquad\qquad \citep{NEURIPS2024_013cf29a} & Designed for drag coefficient prediction from meshes under fixed flow conditions. \\
\bottomrule
\end{tabular}
\end{table}
\section{Evaluation Metrics}
\label{app:metrics}
The four evaluation metrics are defined as follows. MSE and MAE measure absolute prediction errors, with MSE being more sensitive to large deviations due to the squared term. REL-L1 and REL-L2 are scale‑normalized relative errors, which enable fair comparisons across samples with different pressure magnitudes.

\begin{align}
\mathrm{MSE} &= \frac{1}{N}\sum_{i=1}^{N}(\hat{C}_{p,i} - C_{p,i})^2, \\
\mathrm{MAE}  &= \frac{1}{N}\sum_{i=1}^{N}|\hat{C}_{p,i} - C_{p,i}|, \\
\mathrm{REL\text{-}L1} &= \frac{\sum_{i=1}^{N}|\hat{C}_{p,i} - C_{p,i}|}{\sum_{i=1}^{N}|C_{p,i}|}, \\
\mathrm{REL\text{-}L2} &= \frac{\sqrt{\sum_{i=1}^{N}(\hat{C}_{p,i} - C_{p,i})^2}}{\sqrt{\sum_{i=1}^{N}C_{p,i}^2}}.
\end{align}

\section{Training Settings}
\label{app:training}
All models (including FSAN and the baselines) are trained with identical hyperparameters and hardware to ensure fair comparison. The key settings are as follows:

\begin{itemize}
    \setlength{\itemsep}{0pt}
    \item \textbf{Optimizer}: AdamW with $\beta_1=0.9$, $\beta_2=0.999$, weight decay $1\times10^{-4}$, and an initial learning rate of $1\times10^{-4}$.
    \item \textbf{Learning rate schedule}: Cosine annealing with a minimum learning rate of $1\times10^{-6}$ over $50$ epochs, preceded by $3000$ warmup iterations.
    \item \textbf{Loss function}: Mean absolute error (MAE), defined as $\frac{1}{N}\sum_{i=1}^{N}|\hat{C}_{p,i} - C_{p,i}|$.
    \item \textbf{Gradient clipping}: Max norm $1.0$.
    \item \textbf{Epochs}: $50$ epochs for all experiments.
    \item \textbf{Batch size}: $1$ per GPU, resulting in a global batch size of $8$ when using $8$ GPUs.
    \item \textbf{Hardware}: Distributed training on $8$ NVIDIA RTX $4080$ SUPER GPUs (each with $16$ GB memory) using PyTorch's distributed data parallel.
    \item \textbf{Point sampling}: During testing, each sample is uniformly downsampled to 8,192 points using the same protocol as training. The reported mean and standard deviation are computed over three independent training and testing runs with different random seeds.
\end{itemize}

All experiments were repeated three times with different random seeds, and the reported results are the mean values along with the corresponding standard deviations.

\section{Additional Results}
\label{app:additional_results}

\subsection{Comparison with AB-UPT}
\label{app:abupt}
The Emmi-Wing paper \citep{paischer2025going} adopts AB-UPT \citep{alkin2025ab} as a baseline. AB-UPT \citep{alkin2025ab} requires additional volume field data (approximately 5 TB), so a full comparison under our standard protocol is computationally prohibitive. We therefore compare FSAN and AB-UPT on the first 1,000 samples of Emmi-Wing (800 for training, 200 for testing), evaluating normalized surface pressure.

We follow the training configuration described in the Emmi-Wing paper:
Lion optimizer with learning rate $1\times10^{-5}$, linear warmup for the
first 5\% of steps and cosine decay afterwards, float16 precision, weight
decay $\lambda=0.05$, and gradient clipping at 0.25. Inputs are z-score
normalized. For AB-UPT, the hidden dimension is 192, and the number of anchor
points for both volume and surface branches is 16,384. The geometry is
subsampled to 65,536 points and then supernode-pooled to 16,384 points. Since
we use a smaller subset than the original paper, we increase the number of
epochs to 200 to ensure convergence. To match parameter counts, FSAN is
reduced to 15 layers (approximately 7.9M parameters) and trained under the
same configuration as AB-UPT.

\begin{table}[ht]
\small
\centering
\caption{Comparison with AB-UPT on the first 1,000 Emmi-Wing samples.}
\label{tab:abupt}
\begin{tabular}{@{}lccccc@{}}
\toprule
Model & Params & MSE$\downarrow$ & MAE$\downarrow$ & Rel L2$\downarrow$ & Rel L1$\downarrow$ \\
\midrule
AB-UPT & 8.9M & 0.0113 & 0.0434 & 0.0762 & 0.0568 \\
FSAN & 7.9M & 0.0102 & 0.0421 & 0.0747 & 0.0542 \\
\bottomrule
\end{tabular}
\end{table}

FSAN achieves slightly better performance than AB-UPT while using only
surface information, whereas AB-UPT additionally requires volume data. We
therefore draw a conservative conclusion: under comparable performance, FSAN
incurs significantly lower training data cost.

\subsection{Effect of Number of Flow States}
\label{app:num_states}

\begin{figure}[ht]
    \centering
    \includegraphics[width=0.65\linewidth]{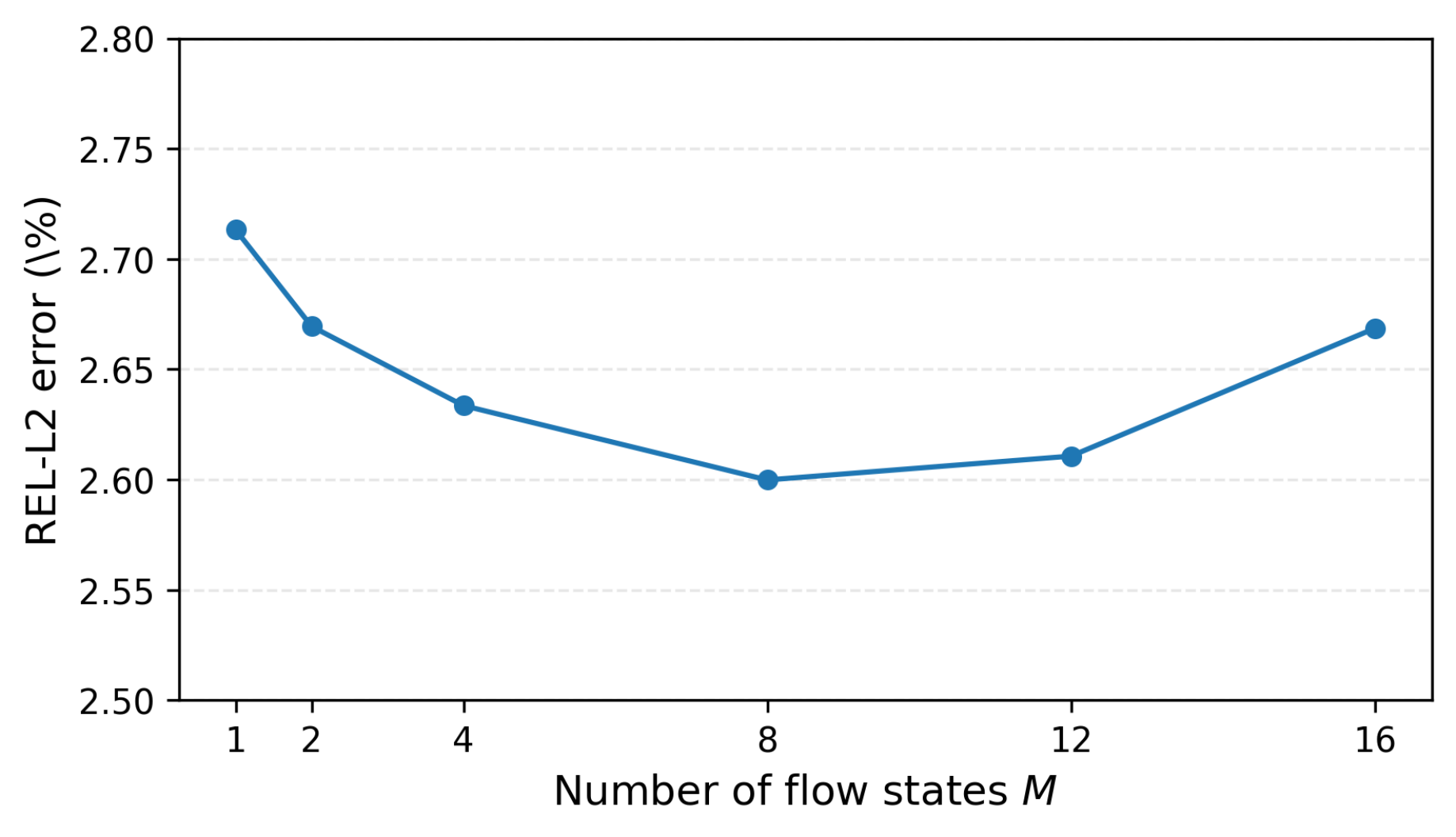}
    \caption{Effect of the number of flow states \(M\) on REL-L2 error.}
    \label{fig:ablation_M}
\end{figure}

We investigate the effect of the number of flow states \(M\) on prediction
accuracy, testing \(M \in \{1, 2, 4, 8, 12, 16\}\). Figure~\ref{fig:ablation_M} reports the REL-L2 error for each configuration.
Accuracy improves as \(M\) increases from 1 to 8, reaching the lowest error
at \(M=8\), and then degrades for \(M=12\) and \(M=16\). This
suggests that over-partitioning may introduce unnecessary complexity.
Therefore, \(M=8\) is used in our full model.

\subsection{Conditioning Strategy Comparison on Transolver}
\label{app:transolver_conditioning}

To validate our baseline adaptation protocol, we compare three conditioning
strategies on the Transolver backbone: concatenation (the protocol used for
geometry-only baselines in this work), FiLM, and AdaLN-Zero. All variants are
trained and evaluated under the same protocol on Emmi-Wing.

\begin{table}[ht]
\small
\centering
\caption{Comparison of conditioning strategies on Transolver (Emmi-Wing).}
\label{tab:transolver_conditioning}
\begin{tabular}{@{}lcccc@{}}
\toprule
Conditioning & MSE$\downarrow$ & MAE$\downarrow$ & REL-L1$\downarrow$ & REL-L2$\downarrow$ \\
             & ($\times 10^{-3}$) & ($\times 10^{-2}$) & (\%) & (\%) \\
\midrule
\textbf{Concatenation} & \textbf{0.375} & \textbf{0.517} & \textbf{1.58} & \textbf{3.24} \\
FiLM          & 0.452 & 0.613 & 1.88 & 3.68 \\
AdaLN-Zero    & 1.047 & 1.048 & 3.23 & 6.44 \\
\bottomrule
\end{tabular}
\end{table}

Table~\ref{tab:transolver_conditioning} shows that concatenation achieves the
best accuracy among the three strategies on this task. This supports our
baseline adaptation protocol: concatenating flow conditions to point
coordinates is a reasonable and effective choice for geometry-only models
that do not natively support multi-parameter flow inputs.

\subsection{Computational Cost Analysis}
\label{app:cost_analysis}

We provide a detailed computational cost analysis of FSAN and the baselines.
The overall cost on Emmi‑Wing is reported in the main text
(Table~\ref{tab:cost}). Here we supplement the cost on DrivAerNet++,
the efficiency under different sampling densities, and a module‑level
breakdown of FSAN's inference cost.

\subsubsection{Overall Computational Cost on DrivAerNet++}
\label{app:cost_drivaer}

Table~\ref{tab:cost_drivaer} reports the computational cost measured on
DrivAerNet++ under the same hardware and sampling settings as Emmi‑Wing
(8$\times$ NVIDIA RTX 4080 SUPER, per‑GPU batch size 1, 8,192 points per
sample). The architecture of FSAN differs only marginally between the two
datasets---the flow condition encoder receives a 3‑dimensional input on
Emmi‑Wing versus a 1‑dimensional input on DrivAerNet++---while the point
cloud processing modules remain identical. The minor differences in the
reported times arise from differences in how the two datasets are loaded.

\begin{table}[htbp]
\small
\centering
\caption{Computational cost comparison on DrivAerNet++.}
\label{tab:cost_drivaer}
\begin{tabular}{@{}lcccccc@{}}
\toprule
Model & Params & Training Time & Inference Time & Train Peak Mem. & Infer. Peak Mem. \\
     & (M)   & (s / batch)  & (s / batch)  & (GB)            & (GB) \\
\midrule
PointTransformerV3 & 46.16 & 0.172  & 0.053  & 2.17 & 0.47 \\
Transolver         & 18.34 & 0.123  & 0.038  & 3.85 & 0.17 \\
AdaField           & 247.08 & 0.977  & 0.465  & 8.63 & 1.34 \\
FSAN (Ours)        & 15.74 & 0.338  & 0.146  & 11.53 & 0.99 \\
\bottomrule
\end{tabular}
\end{table}

\subsubsection{Runtime and Memory across Different Point Densities}
\label{app:point_density}

To analyse the effect of the number of sampled points on efficiency, we
measure the inference time and peak GPU memory of each method on the same
test sample (full surface 370,333 points) at several point
counts. Table~\ref{tab:runtime_memory} reports the per‑batch inference time
and peak memory for each point count, together with the full‑surface
end‑to‑end inference time. ``/'' indicates that the configuration exceeds
the 16~GB memory limit of a single GPU. Transolver can process the full surface in one batch, while FSAN, AdaField, PointTransformerV3, and AB‑UPT require multiple batches.

\begin{table}[ht]
\small
\centering
\caption{Inference time and peak memory at different point counts
(measured on case\_id=27).}
\label{tab:runtime_memory}
\resizebox{\linewidth}{!}{
\begin{tabular}{@{}lccccccc@{}}
\toprule
\multirow{2}{*}{Model} & 8,192 pts & 16,384 pts & 32,768 pts & 65,536 pts & 131,072 pts & 262,144 pts & Full surface \\
 & (ms / GB) & (ms / GB) & (ms / GB) & (ms / GB) & (ms / GB) & (ms / GB) & (s / batches) \\
\midrule
PointTransformerV3 & 49.94 / 0.47 & 51.43 / 0.72 & 57.27 / 1.26 & 70.99 / 2.34 & 100.35 / 4.50 & 161.59 / 8.82 & 0.31 / 3 \\
Transolver & 36.73 / 0.17 & 70.04 / 0.25 & 141.55 / 0.42 & 291.61 / 0.77 & 580.76 / 1.46 & 1171.78 / 2.84 & 1.67 / 1 \\
AdaField & 453.82 / 1.34 & 579.00 / 1.96 & 1004.70 / 5.01 & / & / & / & 11.57 / 12 \\
FSAN (Ours) & 152.90 / 0.99 & 396.81 / 1.86 & 1182.81 / 4.14 & / & / & / & 13.22 / 12 \\
\bottomrule
\end{tabular}
}
\end{table}

FSAN shows a steeper increase in inference time and memory with the number
of points than the hierarchical baselines, mainly because its non‑hierarchical
local attention retains full‑resolution activations. The full‑surface
end‑to‑end time is correspondingly larger, as FSAN requires 12 chunks to
cover the complete surface, while Transolver needs only one.

\subsubsection{Module-level Cost Breakdown}
\label{app:module_cost}

We further measure the contribution of each FSAN module to inference time
and peak GPU memory. Inference time is recorded with
\texttt{torch.cuda.synchronize} and \texttt{time.perf\_counter} at module
boundaries. Peak memory contribution is measured through differential
experiments: we build the full model and variants where a specific module is
removed, and compare the peak memory. Table~\ref{tab:module_cost} reports the
time fraction and peak memory fraction of each module at 8,192, 16,384, and
32,768 points.

\begin{table}[ht]
\small
\centering
\caption{Module-level time and peak memory fractions for FSAN at different
point counts. In each cell, the left value is the time fraction and the right
value is the peak memory fraction.}
\label{tab:module_cost}
\begin{tabular}{@{}lccc@{}}
\toprule
Module & 8192 pts & 16384 pts & 32768 pts \\
\midrule
Local attention layer & 90.1\% / 81.4\% & 95.4\% / 87.2\% & 97.6\% / 91.5\% \\
Flow condition encoder & 1.9\% / 0.0\% & 0.8\% / 0.0\% & 0.3\% / 0.0\% \\
Flow state attention & 3.2\% / 0.0\% & 1.3\% / 0.0\% & 0.4\% / 0.0\% \\
Others & 4.8\% / 18.6\% & 2.5\% / 12.8\% & 1.6\% / 8.5\% \\
\bottomrule
\end{tabular}
\end{table}

The local attention layer dominates both inference time and peak memory. In contrast, the proposed flow state attention module accounts for less than 3.2\% of inference time and has a negligible memory footprint. This confirms that the computational overhead of FSAN stems from the non‑hierarchical point cloud backbone, while the core contribution---flow state attention---is lightweight.

\subsection{Sensitivity to Sampling Density}
\label{app:sensitivity}

We evaluate FSAN on the full test set under different numbers of sampled
points to study its sensitivity to sampling density. The model is trained
with 8,192 points and is tested without further fine-tuning. For each
configuration, we randomly downsample each surface to the specified number
of points and compute MSE, MAE, REL-L1, and REL-L2.

\begin{table}[ht]
\small
\centering
\caption{Sensitivity of FSAN to the number of sampled points on Emmi-Wing.}
\label{tab:sensitivity}
\begin{tabular}{@{}lcccc@{}}
\toprule
Points & MSE$\downarrow$ & MAE$\downarrow$ & REL-L1$\downarrow$ & REL-L2$\downarrow$ \\
       & ($\times 10^{-3}$) & ($\times 10^{-2}$) & (\%) & (\%) \\
\midrule
1,024   & 1.927 & 2.417 & 7.77 & 9.62 \\
2,048   & 0.747 & 1.216 & 3.86 & 5.46 \\
4,096   & 0.365 & 0.607 & 1.88 & 3.18 \\
8,192   & \textbf{0.297} & \textbf{0.456} & \textbf{1.40} & \textbf{2.59} \\
16,384  & 0.330 & 0.566 & 1.75 & 2.93 \\
32,768  & 0.482 & 0.836 & 2.62 & 4.26 \\
\bottomrule
\end{tabular}
\end{table}

As shown in Table~\ref{tab:sensitivity}, FSAN achieves the best accuracy at
8,192 points, which matches the training density. Reducing the number of
points below 4,096 leads to a noticeable degradation in accuracy, while
increasing the number of points beyond 8,192 also slightly degrades
performance. We attribute this to the mismatch between training and testing
point densities.

\subsection{Consistency between Full-Surface and Sampled Predictions}
\label{app:full_surface}
We compare full-surface predictions with standard 8,192-point sampled
predictions to check whether the random partition used during full-surface
inference introduces any boundary artifacts. For each test sample, the full
point cloud is randomly shuffled and sequentially split into chunks of
8,192 points; the last two chunks are merged to avoid undersized final
batches. Each chunk is processed independently, and the predictions are then
stitched back to the original point order. The evaluation is performed on
all test samples of Emmi-Wing.

\begin{table}[ht]
\small
\centering
\caption{Full-surface prediction vs.\ 8,192-point sampled prediction on
Emmi-Wing.}
\label{tab:full_surface}
\begin{tabular}{@{}lcccc@{}}
\toprule
Evaluation & MSE$\downarrow$ & MAE$\downarrow$ & REL-L1$\downarrow$ & REL-L2$\downarrow$ \\
           & ($\times 10^{-3}$) & ($\times 10^{-2}$) & (\%) & (\%) \\
\midrule
Sampled (8,192 pts) & 0.297 & 0.456 & 1.40 & 2.59 \\
Full-surface (stitched) & 0.298 & 0.457 & 1.40 & 2.63 \\
\bottomrule
\end{tabular}
\end{table}

The full-surface results are nearly identical to the sampled predictions,
with only a marginal difference in REL-L2 (2.63\% vs.\ 2.59\%). This
indicates that the randomized chunking and stitching procedure introduces
negligible boundary errors, and that the model behaves consistently for
points processed in different batches.

\subsection{Qualitative Results}
\label{app:qualitative}

\subsubsection{Prediction Quality}

We visualise the predicted pressure coefficient fields and the corresponding
signed prediction errors for four representative test samples: a moderate
case, a high angle-of-attack case, a high Mach and Reynolds number case,
and an extreme case combining all three factors. For each sample, the full
surface is processed by random chunking with 8,192 points per chunk, as
described in Appendix~\ref{app:full_surface}. For visual clarity, the error
colourmap is saturated at \(\pm 0.015\); signed errors beyond this range are
clipped to \([-0.015, 0.015]\). The corresponding unclipped error
statistics---the percentage of points exceeding the saturation threshold,
the maximum absolute error, and the 95th percentile absolute error---are
summarised in Table~\ref{tab:error_stats}.

\begin{figure}
    \centering
    \includegraphics[width=\linewidth]{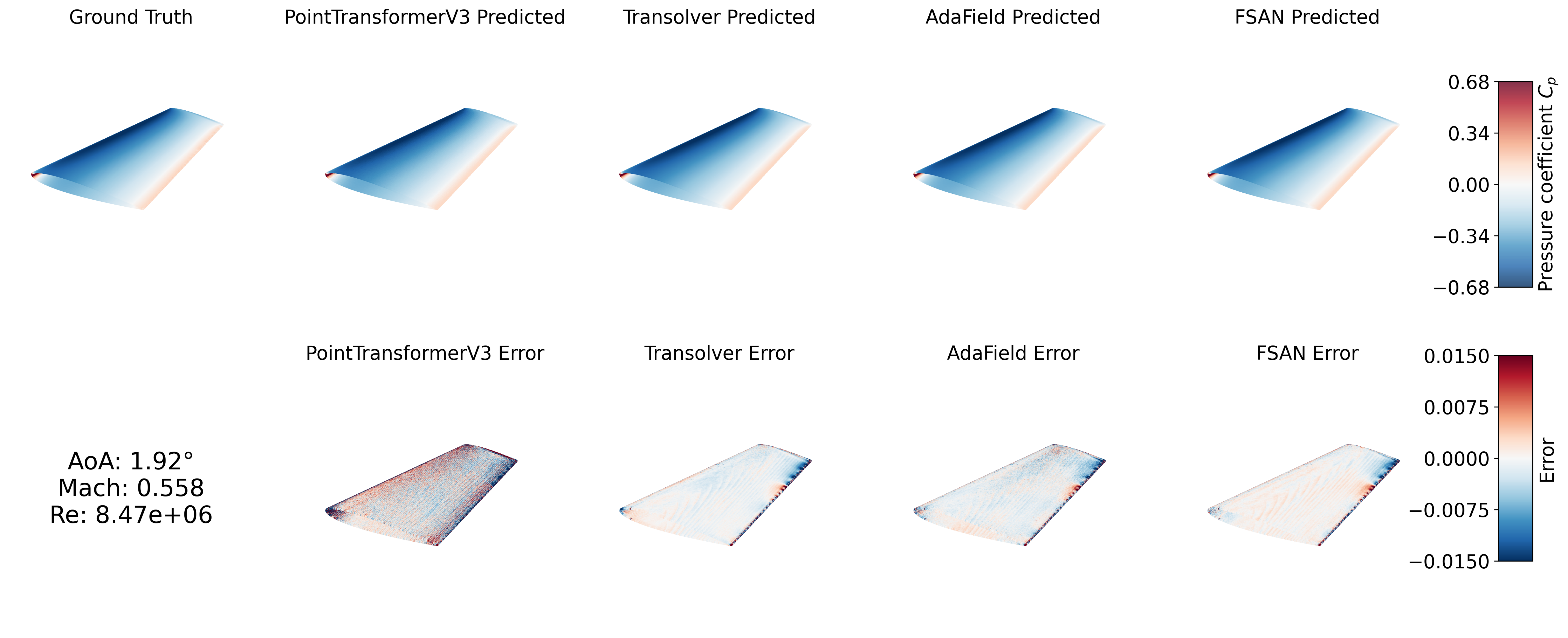}
    \caption{Moderate condition (AoA \(= 1.92^\circ\), Mach \(= 0.558\),
    Re \(= 8.47\times10^{6}\)). Top: predicted \(C_p\). Bottom: signed
    prediction error, saturated at \(\pm 0.015\). Unclipped statistics are
    reported in Table~\ref{tab:error_stats}.}
    \label{fig:comp_moderate}
\end{figure}

\begin{figure}
    \centering
    \includegraphics[width=\linewidth]{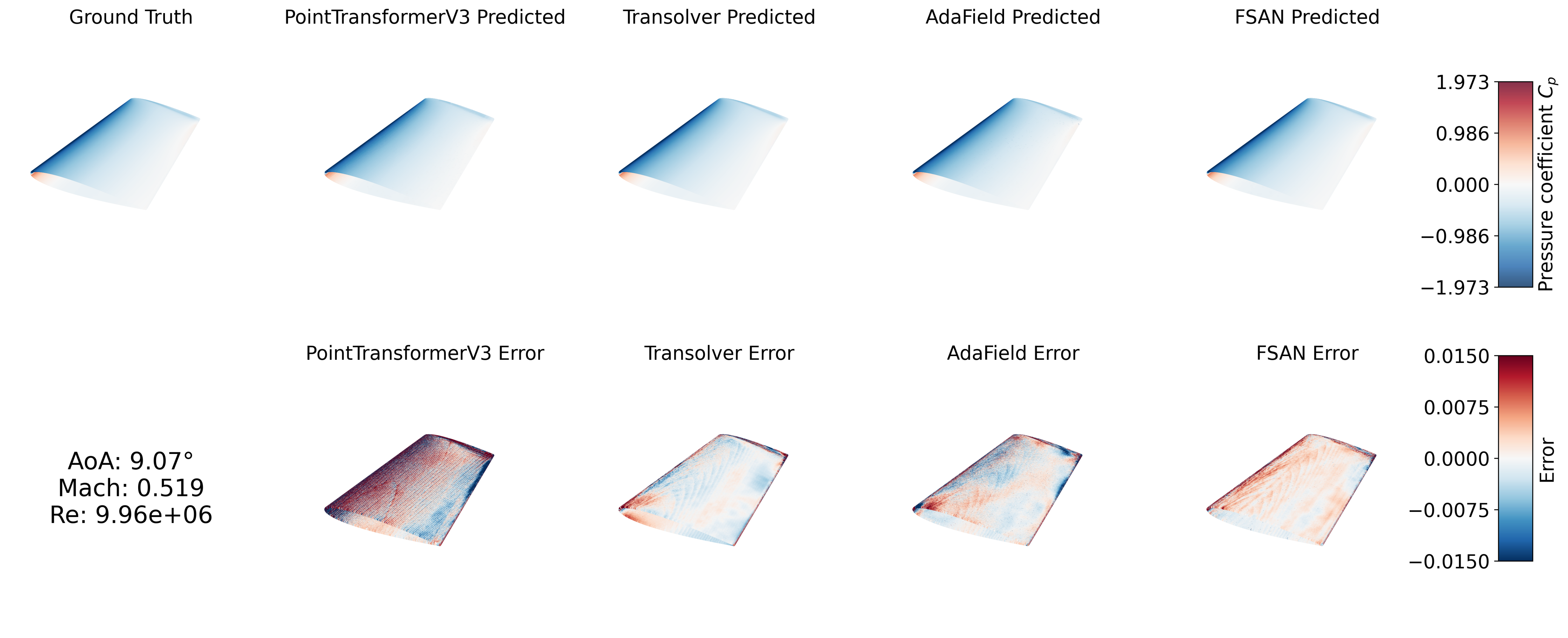}
    \caption{High angle of attack (AoA \(= 9.07^\circ\), Mach \(= 0.519\),
    Re \(= 9.96\times10^{6}\)). Top: predicted \(C_p\). Bottom: signed
    prediction error, saturated at \(\pm 0.015\). Unclipped statistics are
    reported in Table~\ref{tab:error_stats}.}
    \label{fig:comp_high_aoa}
\end{figure}

\begin{figure}
    \centering
    \includegraphics[width=\linewidth]{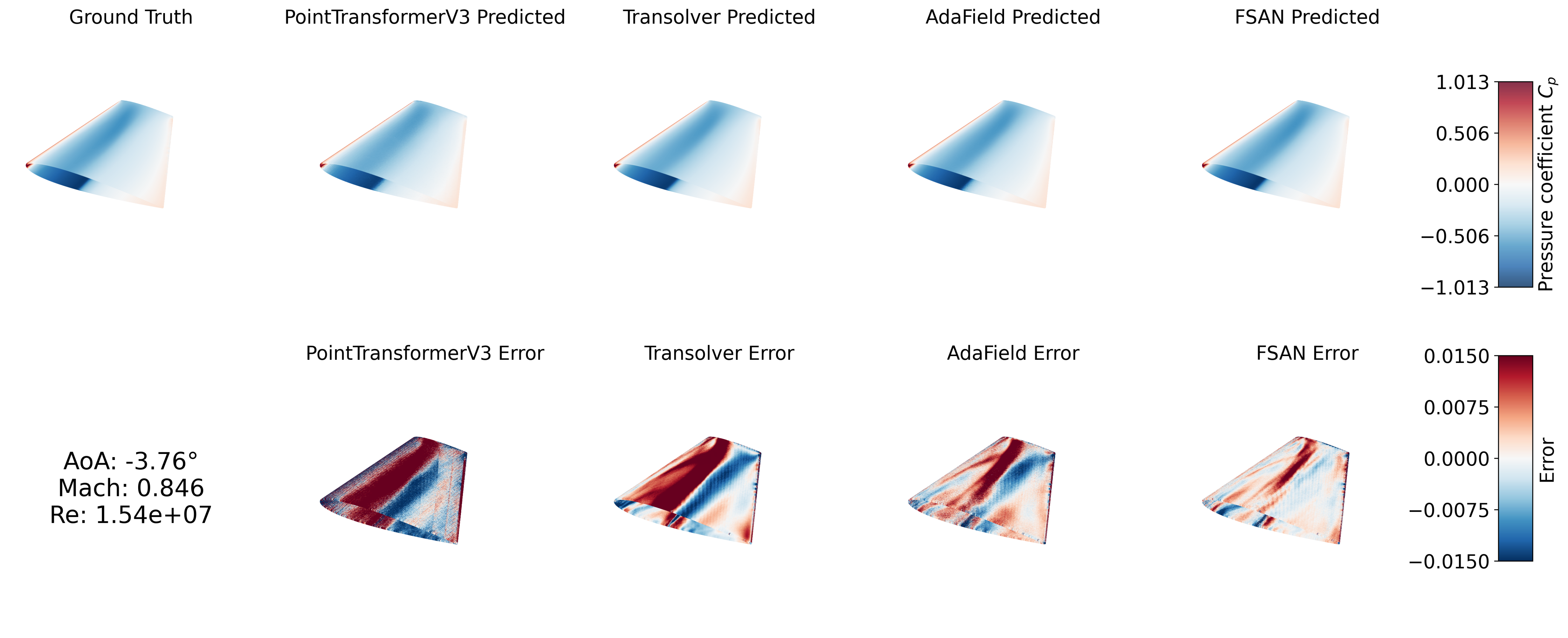}
    \caption{High Mach and Reynolds numbers (AoA \(= -3.76^\circ\),
    Mach \(= 0.846\), Re \(= 1.54\times10^{7}\)). Top: predicted \(C_p\).
    Bottom: signed prediction error, saturated at \(\pm 0.015\). Unclipped
    statistics are reported in Table~\ref{tab:error_stats}.}
    \label{fig:comp_high_mach}
\end{figure}

\begin{figure}[ht]
    \centering
    \includegraphics[width=\linewidth]{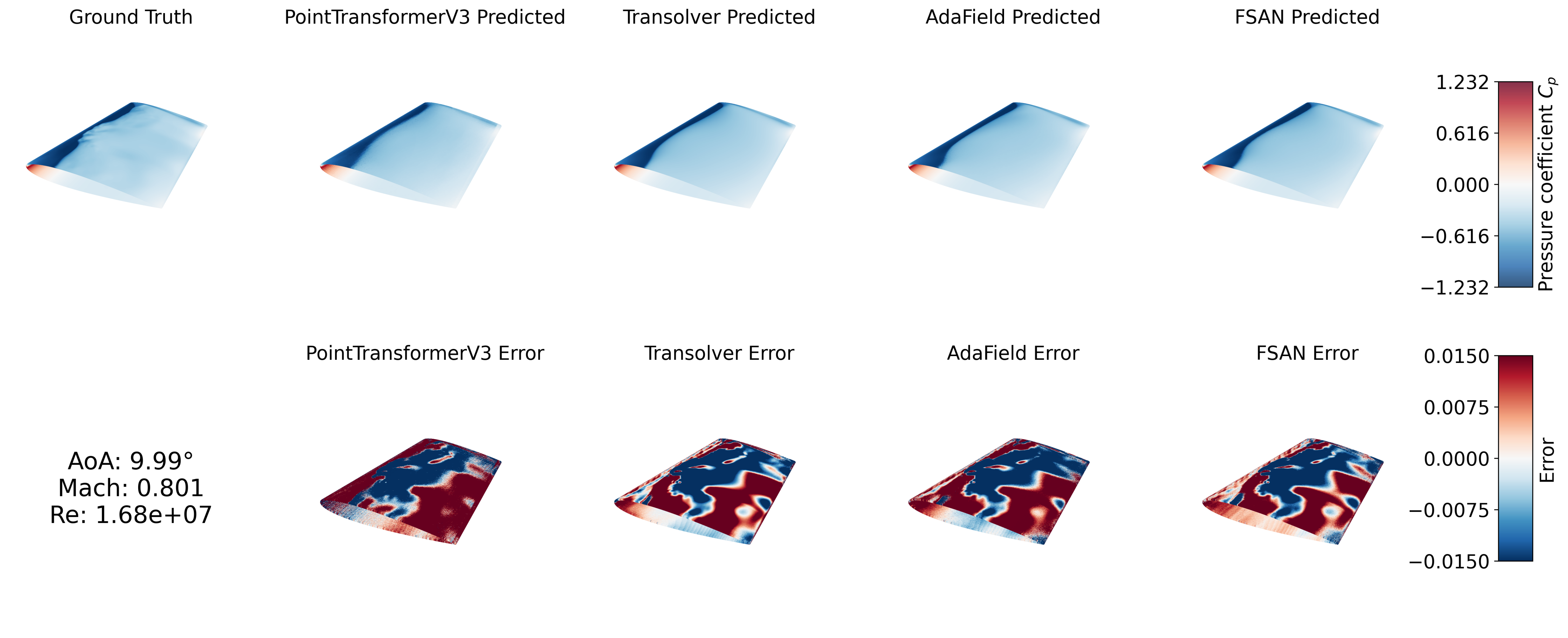}
    \caption{Extreme in-distribution condition (AoA \(= 9.99^\circ\),
    Mach \(= 0.801\), Re \(= 1.68\times10^{7}\)). Top: predicted \(C_p\).
    Bottom: signed prediction error, saturated at \(\pm 0.015\). Unclipped
    statistics are reported in Table~\ref{tab:error_stats}.}
    \label{fig:comp_extreme}
\end{figure}

As shown in Figures~\ref{fig:comp_moderate}--\ref{fig:comp_high_mach}, FSAN
achieves visibly lower errors than the baselines on the first three samples.
Under the extreme combined condition (Figure~\ref{fig:comp_extreme}), all
methods exhibit larger errors, which suggests that the difficulty reflects
the intrinsic complexity of the flow regime rather than a specific
limitation of FSAN.

\begin{table}[ht]
\setlength{\tabcolsep}{4pt}
\small
\centering
\caption{Unclipped error statistics for the four representative test
samples.}
\label{tab:error_stats}
\begin{tabular}{@{}c l ccc@{}}
\toprule
\multirow{2}{*}{Fig.} & \multirow{2}{*}{Method} & Clipped & Max & P95 \\
 & & (\%) & & \\
\midrule
\multirow{4}{*}{\ref{fig:comp_moderate}}
 & PointTransformerV3  & 8.8  & 2.17 & 0.029 \\
 & Transolver          & 1.0  & 1.38 & 0.006 \\
 & AdaField            & 1.4  & 0.90 & 0.007 \\
 & FSAN (Ours)         & 1.0  & 0.38 & 0.006 \\
\cmidrule{1-5}
\multirow{4}{*}{\ref{fig:comp_high_aoa}}
 & PointTransformerV3  & 18.6 & 2.32 & 0.060 \\
 & Transolver          & 2.5  & 1.91 & 0.009 \\
 & AdaField            & 4.1  & 1.68 & 0.013 \\
 & FSAN (Ours)         & 2.5  & 0.37 & 0.009 \\
\cmidrule{1-5}
\multirow{4}{*}{\ref{fig:comp_high_mach}}
 & PointTransformerV3  & 44.6 & 2.18 & 0.076 \\
 & Transolver          & 16.2 & 2.07 & 0.029 \\
 & AdaField            & 12.4 & 1.47 & 0.025 \\
 & FSAN (Ours)         & 3.6  & 1.20 & 0.013 \\
\cmidrule{1-5}
\multirow{4}{*}{\ref{fig:comp_extreme}}
 & PointTransformerV3  & 49.2 & 2.57 & 0.097 \\
 & Transolver          & 31.9 & 1.75 & 0.062 \\
 & AdaField            & 37.0 & 1.25 & 0.066 \\
 & FSAN (Ours)         & 29.9 & 1.47 & 0.059 \\
\bottomrule
\end{tabular}
\end{table}

\subsubsection{Evolution of Flow State Assignments}

To better understand the behaviour of the flow state partition, we visualise
the dominant flow state at each of the 12 FSAN blocks for the four
representative samples. Here, the dominant state at each point is defined as
the state with the highest softmax assignment weight. The state indices are
independent across blocks---State~1 at Block~3, for example, does not
correspond to State~1 at Block~4---so the visualisations should be read layer by layer.

Several patterns emerge from these visualisations. The partition at the
first block is similar across all four samples, consistent with the design
that the initial assignment is driven purely by geometric features. As the
blocks deepen, the partitions diverge across samples, reflecting the
accumulating influence of flow conditions through the cross-attention
updates. In addition, the dominant orientation of the partitions gradually
shifts from spanwise in early blocks to chordwise in later blocks, which is
more directly relevant to the surface pressure distribution. These
observations indicate that the learned states are not fixed geometric
clusters, but are continuously reorganised across blocks in a task-driven
manner.

\begin{figure}
    \centering
    \includegraphics[width=0.9\textwidth]{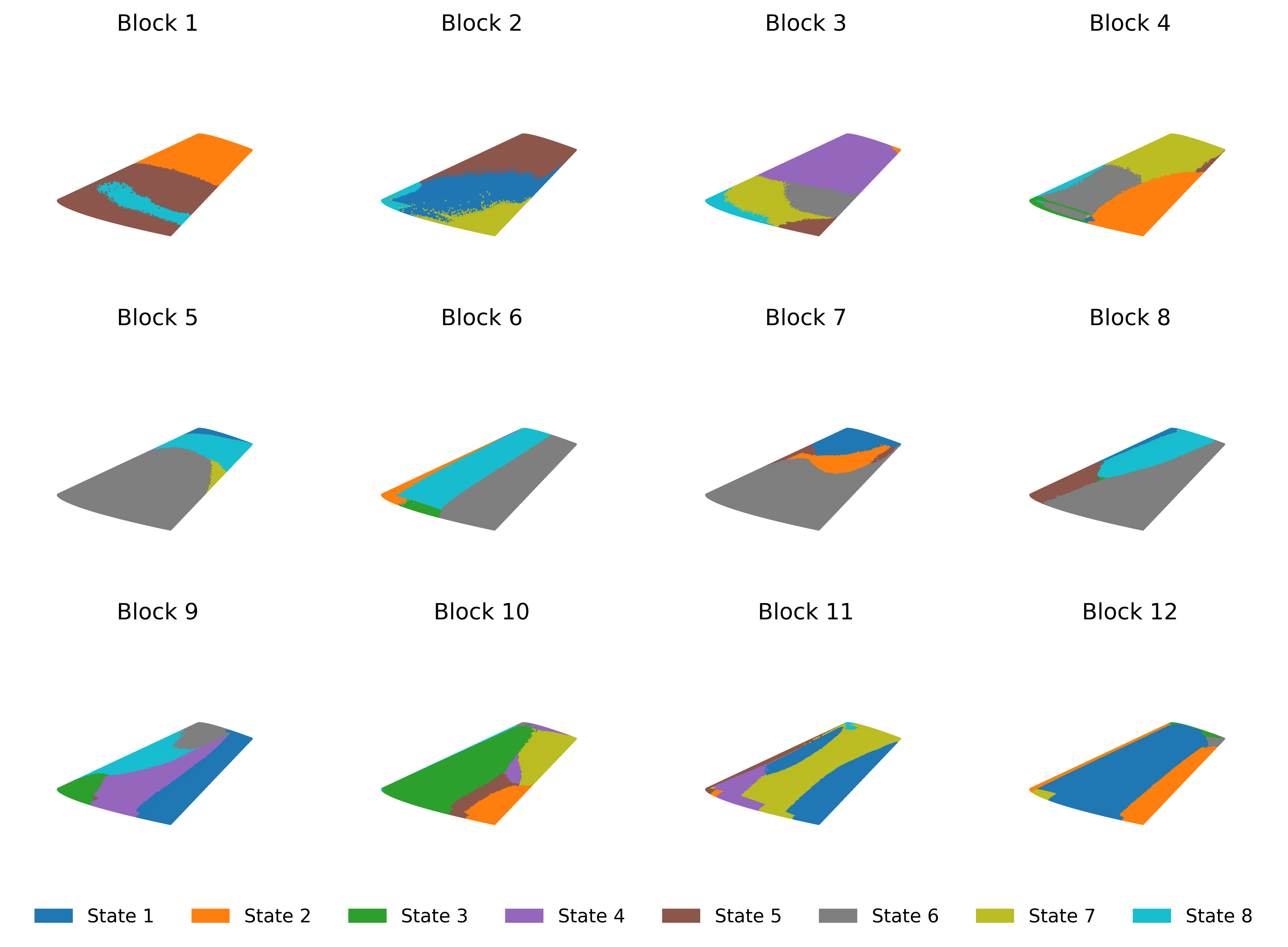}
    \caption{Dominant flow state per block for the moderate sample. State
    indices are independent across blocks.}
    \label{fig:state_prog_1}
\end{figure}

\begin{figure}
    \centering
    \includegraphics[width=0.9\textwidth]{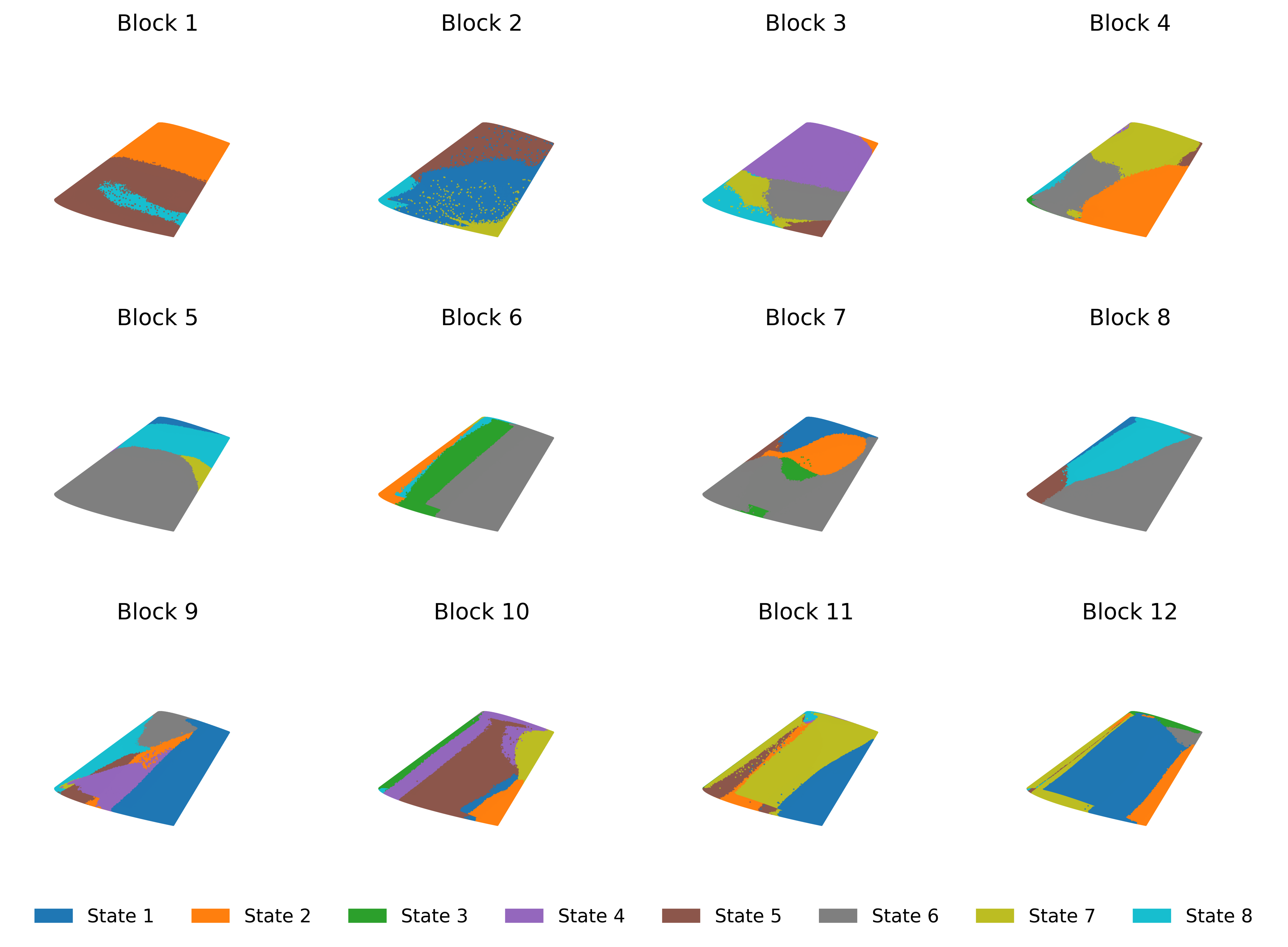}
    \caption{Dominant flow state per block for the high angle-of-attack
    sample. State indices are independent across blocks.}
    \label{fig:state_prog_2}
\end{figure}

\begin{figure}
    \centering
    \includegraphics[width=0.9\textwidth]{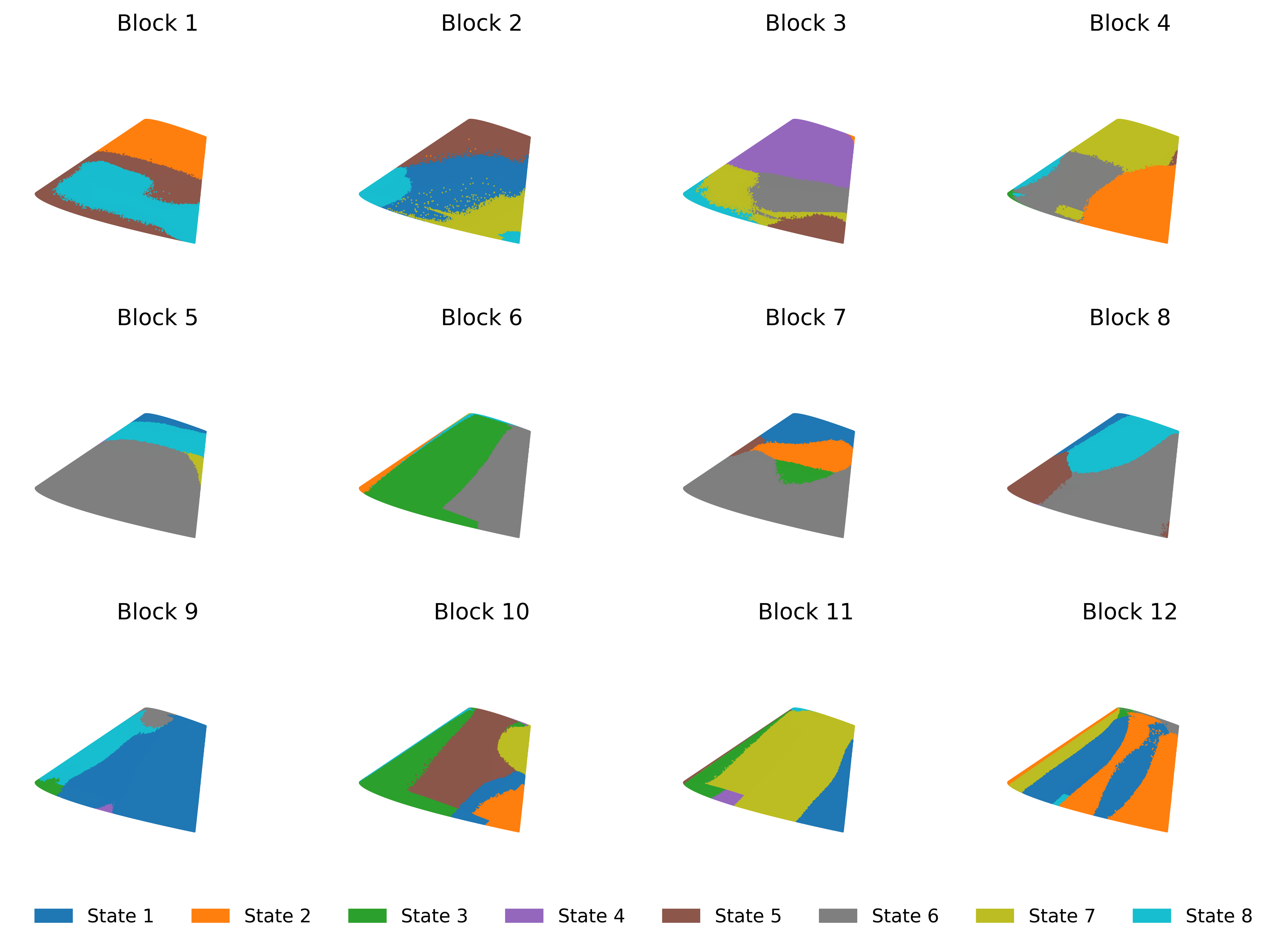}
    \caption{Dominant flow state per block for the high Mach and Reynolds
    number sample. State indices are independent across blocks.}
    \label{fig:state_prog_3}
\end{figure}

\begin{figure}
    \centering
    \includegraphics[width=0.9\textwidth]{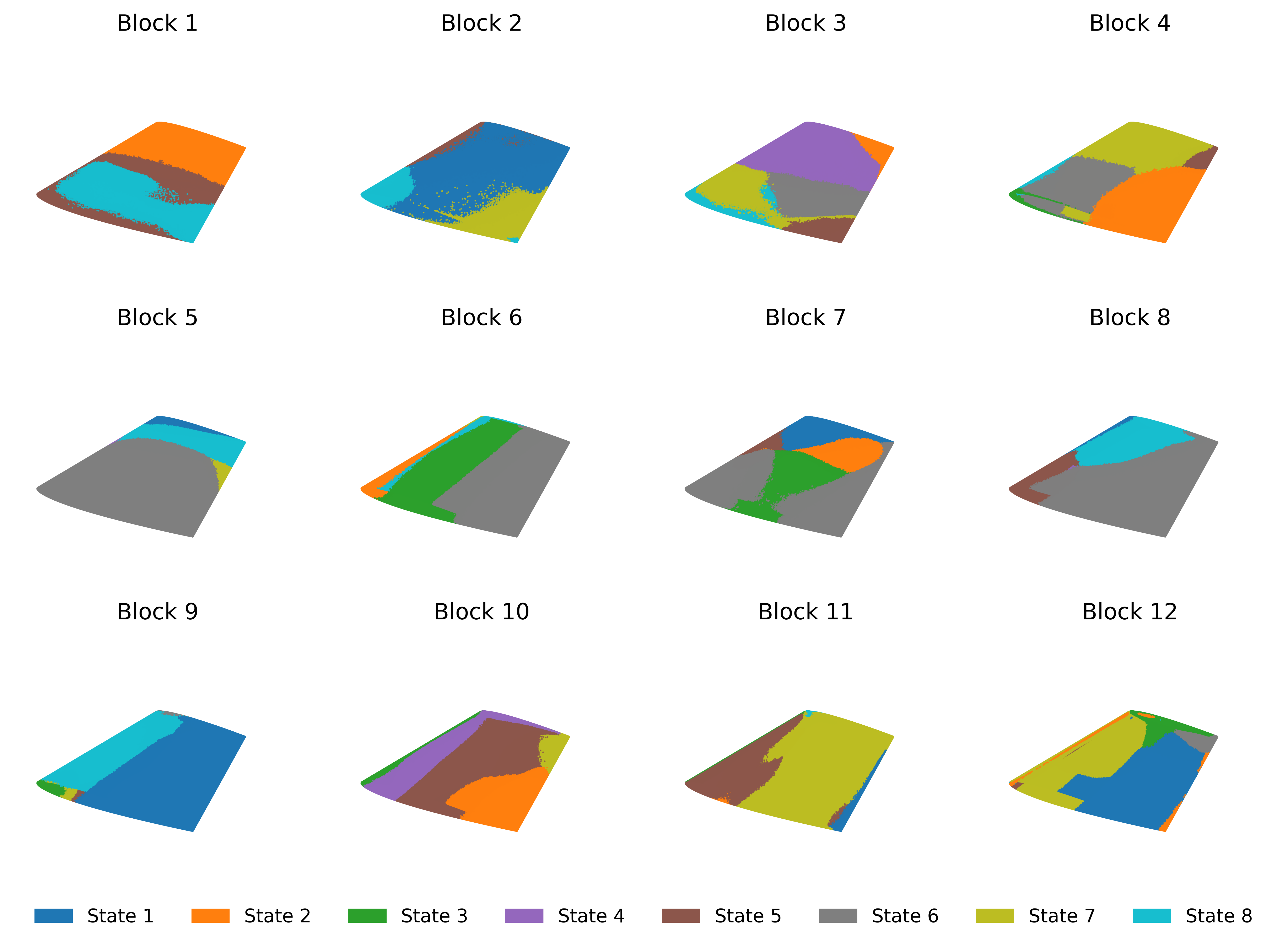}
    \caption{Dominant flow state per block for the extreme combined sample.
    State indices are independent across blocks.}
    \label{fig:state_prog_4}
\end{figure}
\clearpage

\subsubsection{Statistical Association with Shock Regions}

We further examine whether the learned flow states exhibit statistical
associations with known aerodynamic structures. Using three transonic
samples from Emmi-Wing, we approximate shock regions by the points with the
largest pressure gradient magnitudes (top \(1\%\) of \(|\nabla C_p|\)),
since shocks are characterized by sharp localised pressure changes. The
non-shock reference is formed by the \(50\%\) points with the smallest
\(|\nabla C_p|\). Table~\ref{tab:shock_ratios} reports the mean weight ratio
of each flow state in the last block between these two regions.

\begin{figure}[ht]
    \centering
    \includegraphics[width=\linewidth]{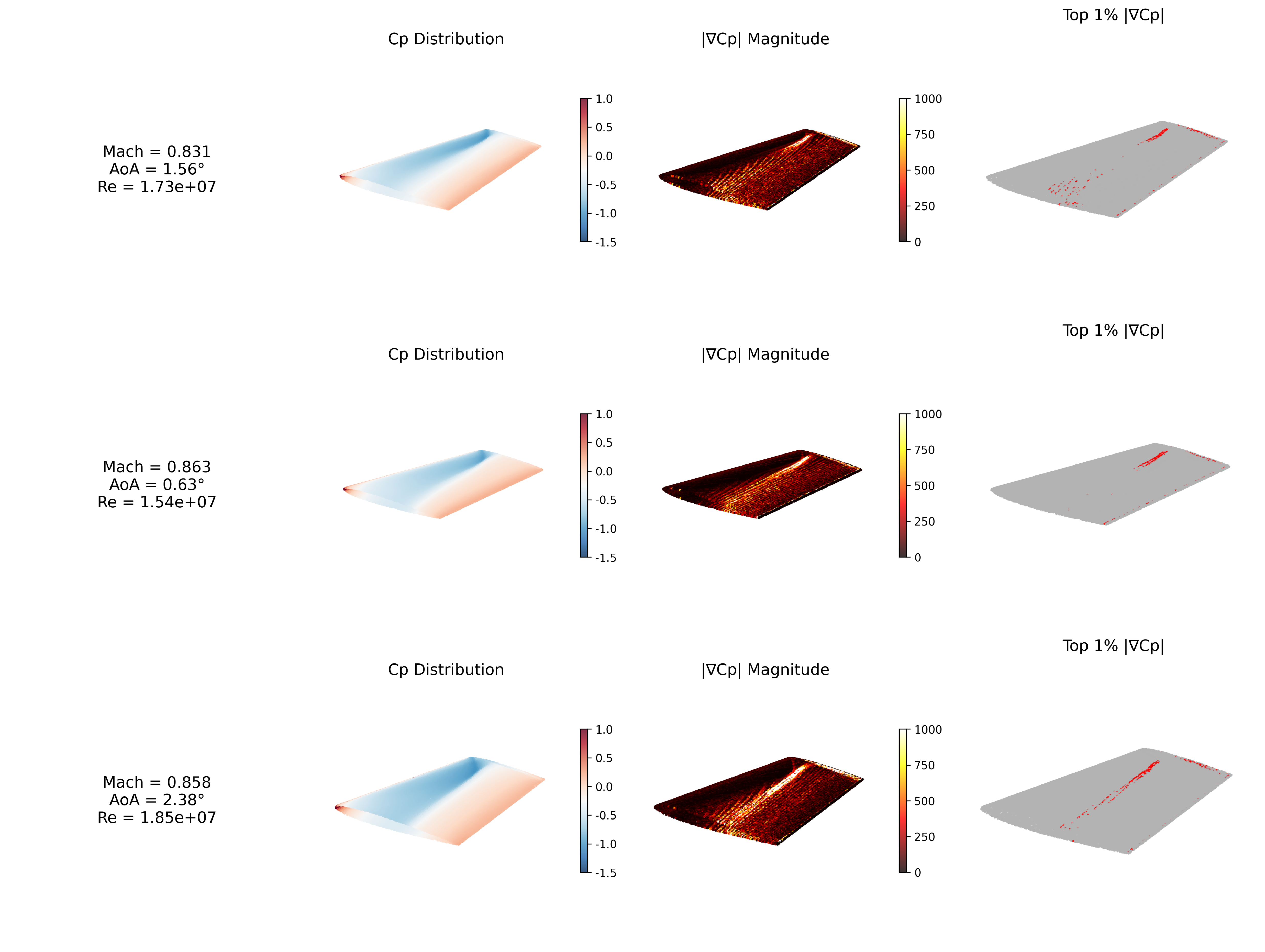}
    \caption{Visualisation of three transonic samples.}
    \label{fig:shock_visualization}
\end{figure}

One flow state consistently shows higher activation on shock regions than
on non-shock regions, while other states avoid shock regions. Since FSAN
uses a softmax-based soft assignment, this observation does not imply a
hard one-to-one mapping to physical flow regimes. Nevertheless, it suggests
that the partition spontaneously captures physically relevant structures in
a data-driven manner. To complement the tabular summary, we also visualise
the pressure gradient magnitude and the spatial distribution of the most
shock-sensitive flow state for the three samples.

\begin{table}[ht]
\small
\centering
\caption{Mean weight ratios (shock / non-shock) for each flow state on
three transonic samples.}
\label{tab:shock_ratios}
\begin{tabular}{@{}c c c c c c c c c c@{}}
\toprule
Case & S1 & S2 & S3 & S4 & S5 & S6 & S7 & S8 \\
\midrule
1  & 1.15 & 1.31 & 1.07 & 0.45 & 0.24 & \textbf{2.17} & 0.15 & 0.78 \\
2  & 1.11 & 1.47 & 1.15 & 0.33 & 0.14 & \textbf{1.97} & 0.09 & 0.58 \\
3 & 1.10 & 1.23 & 1.46 & 0.36 & 0.22 & \textbf{1.95} & 0.12 & 0.81 \\
\bottomrule
\end{tabular}
\end{table}

\end{document}